\documentclass[10pt,twocolumn,letterpaper]{article}
\usepackage{geometry}
\usepackage{authblk}
\usepackage{times}
\usepackage{epsfig}
\usepackage{graphicx}
\usepackage{amsmath}
\usepackage{amssymb}
\usepackage{cuted}
\usepackage[ruled,vlined]{algorithm2e} 
\SetEndCharOfAlgoLine{}
\usepackage{tabularx}
\usepackage{subcaption}

\usepackage[inline]{enumitem}

\usepackage{tikz}
\usetikzlibrary{shapes.geometric,arrows,positioning,automata}
\tikzstyle{startstop} = [rectangle, rounded corners, minimum width=2cm, minimum
height=1cm,text centered, draw=black, fill=red!30]
\tikzstyle{process} = [rectangle, minimum width=2cm, minimum height=1cm, text
centered, draw=black, fill=orange!30]
\tikzstyle{decision} = [diamond, minimum width=2cm, minimum height=1cm, text
centered, draw=black, fill=green!30]
\tikzstyle{io} = [rectangle, minimum width=2cm, minimum height=1cm, text
centered, draw=black, fill=blue!30]
\tikzstyle{arrow} = [thick,->,>=stealth]
\usepackage[pagebackref=true, breaklinks=true, letterpaper=true, colorlinks,
bookmarks=true] {hyperref}
\usepackage{pgfplots}
\pgfplotsset{compat=1.12}

\begin{document}
\title{\vspace{-0.5 in}Real-time Visual Object Tracking with Natural Language Description}
\date{}
% THIS FILE IS RESERVED FOR DEFINITIONS AND AUTHORS.

\author[1]{Qi Feng}
\author[1]{Vitaly Ablavsky}
\author[2]{Qinxun Bai\thanks{Work done at Hikvision Research America}}
\author[3]{Guorong Li}
\author[1]{Stan Sclaroff}
\affil[1]{Department of Computer Science, Boston University}
\affil[2]{Horizon Robotics}
\affil[3]{University of Chinese Academy of Sciences}
\affil[1]{\textit{\{fung,ablavsky,sclaroff\}@bu.edu}}
\affil[2]{\textit{qinxun.bai@gmail.com}}
\affil[3]{\textit{liguorong@ucas.edu.cn}}

\maketitle
\begin{abstract}
  In recent years, deep learning based visual object trackers have been studied
  thoroughly, but handling occlusions and/or rapid motions of the target
  remains challenging. In this work, we argue that conditioning on the natural
  language (NL) description of a target provides information for longer-term
  invariance, and thus helps cope with typical tracking challenges. However,
  deriving a formulation to combine the strengths of appearance-based tracking
  with the language modality is not straightforward. Therefore, we propose a
  novel deep tracking-by-detection formulation that can take advantage of NL
  descriptions. Regions that are related to the given NL description are
  generated by a proposal network during the detection phase of the tracker.
  Our LSTM based tracker then predicts the update of the target from regions
  proposed by the NL based detection phase. Our method runs at over 30 fps on a
  single GPU. In benchmarks, our method is competitive with state of the art
  trackers that employ bounding boxes for initialization, while it outperforms
  all other trackers on targets given unambiguous and precise language
  annotations. When conditioned on only NL descriptions, our model doubles the
  performance of the previous best attempt~\cite{Li}.
\end{abstract}
\section{Introduction}
Progress in visual object tracking has been driven in part by the establishment
of standard datasets and
competitions/challenges~\cite{VOT_TPAMI,yilmaz2006object} tied to common
evaluation metrics~\cite{fan2018lasot,yilmaz2006object}. These datasets range
from specific scenarios, such as video surveillance~\cite{Geiger2013IJRR},
containing predominantly pedestrians and vehicles captured with stationary
cameras, to datasets that depicts objects of unrestricted categories captured
with cameras undergoing arbitrary motion. Most recent approaches focused on the
latter types of datasets, proposing new types of deep
features~\cite{bertinetto2016fully,li2019siamrpn++,li2018high,zhu2018distractor}
that are robust to occlusions, rapid motion, etc.

However, since the ``unrestricted categories'' datasets require good tracking
performance across a wide range of scenarios, and this affects how the tracking
problem itself is posed. Specifically, it has been a  {\em de~facto} standard
to initialize a tracker with a bounding box in the first frame, and require
that this bounding box accurately cover the object of interest. Such a
requirement stands in contrast to tracking in the surveillance domain, where a
tracker is expected to self-initialize using proposals for regions likely to
contain objects of interest~\cite{Geiger2013IJRR}.

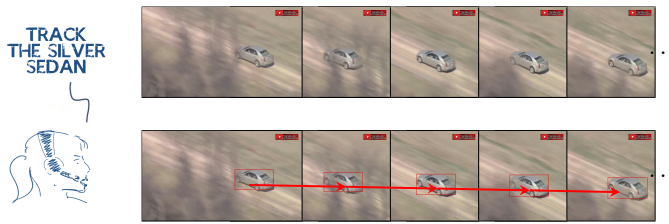
\begin{figure}[t]
  \begin{center}
    \resizebox{\columnwidth}{!}{
      \begin{tikzpicture}
        \node (img5) {\frame{\includegraphics[height=0.5in]{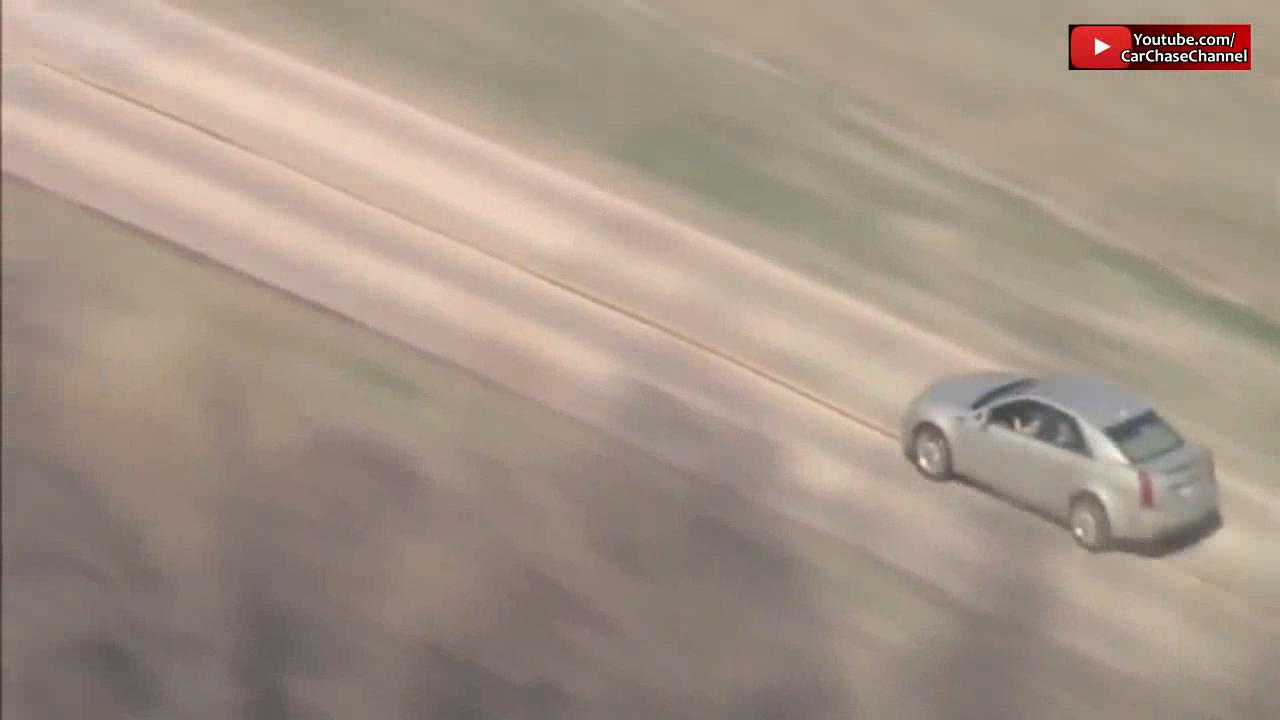}}};
        \node (img4) at (img5.west) {\frame{\includegraphics[height=0.5in]{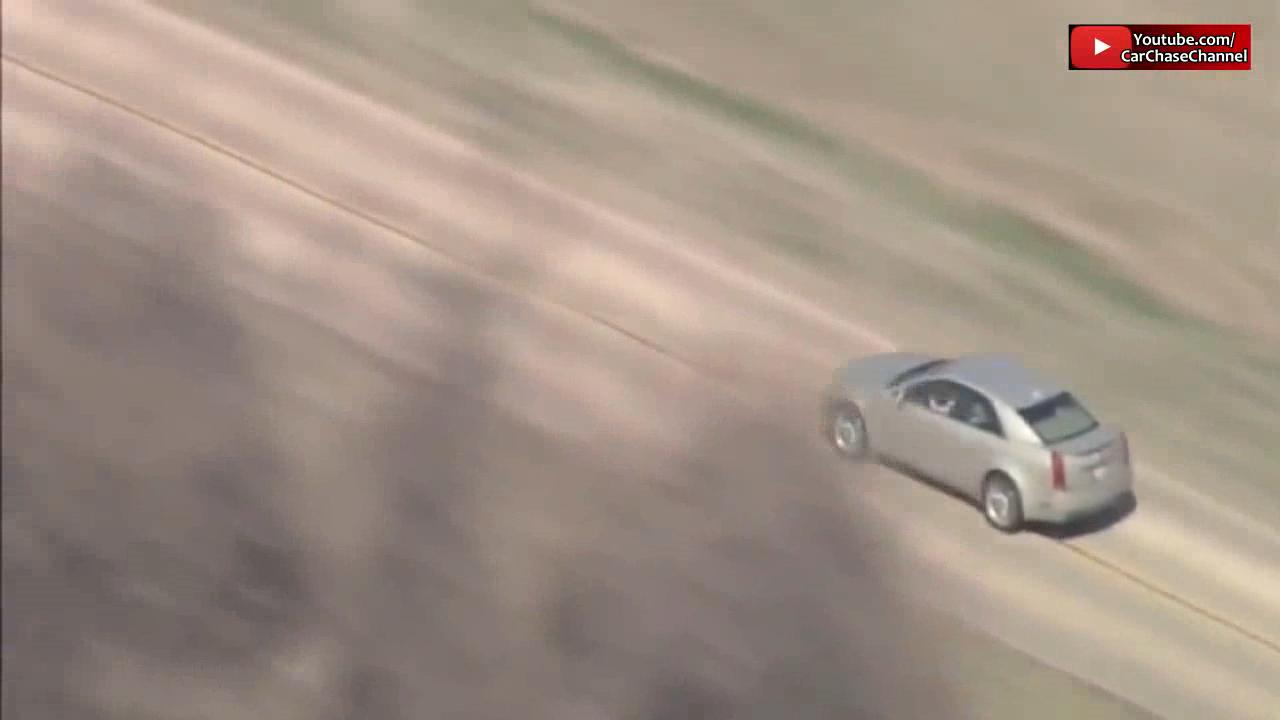}}};
        \node (img3) at (img4.west) {\frame{\includegraphics[height=0.5in]{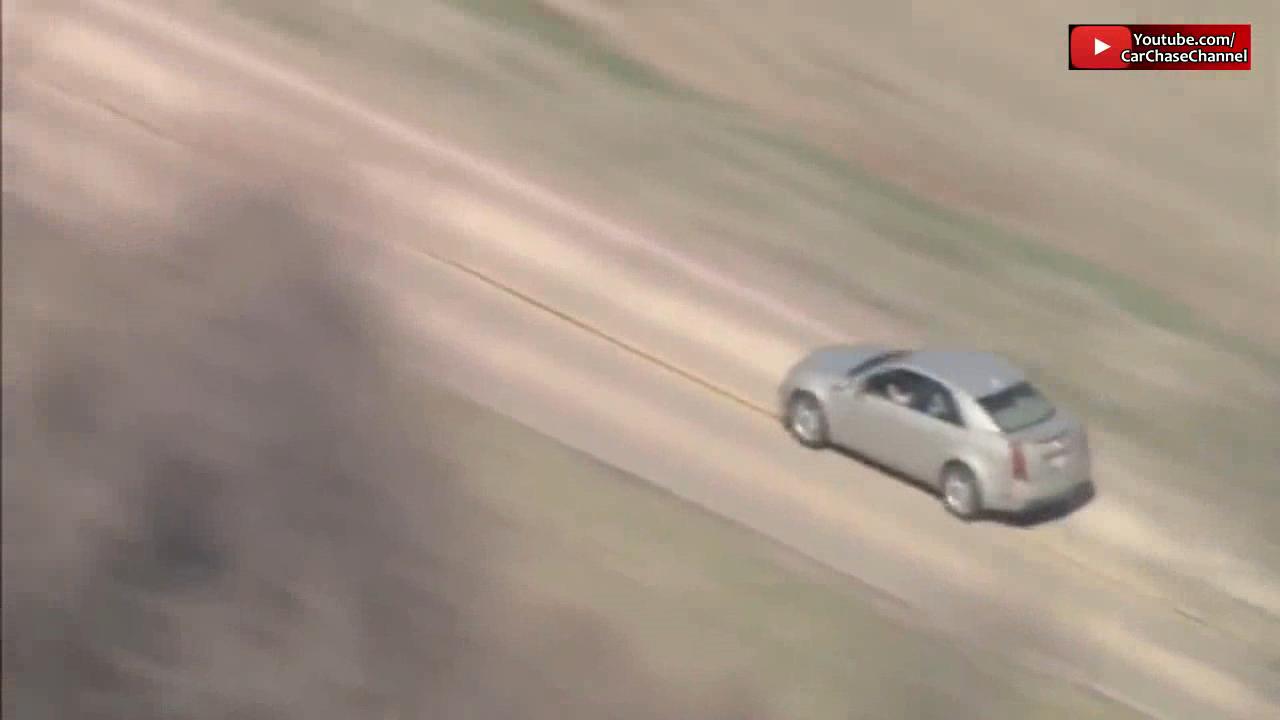}}};
        \node (img2) at (img3.west) {\frame{\includegraphics[height=0.5in]{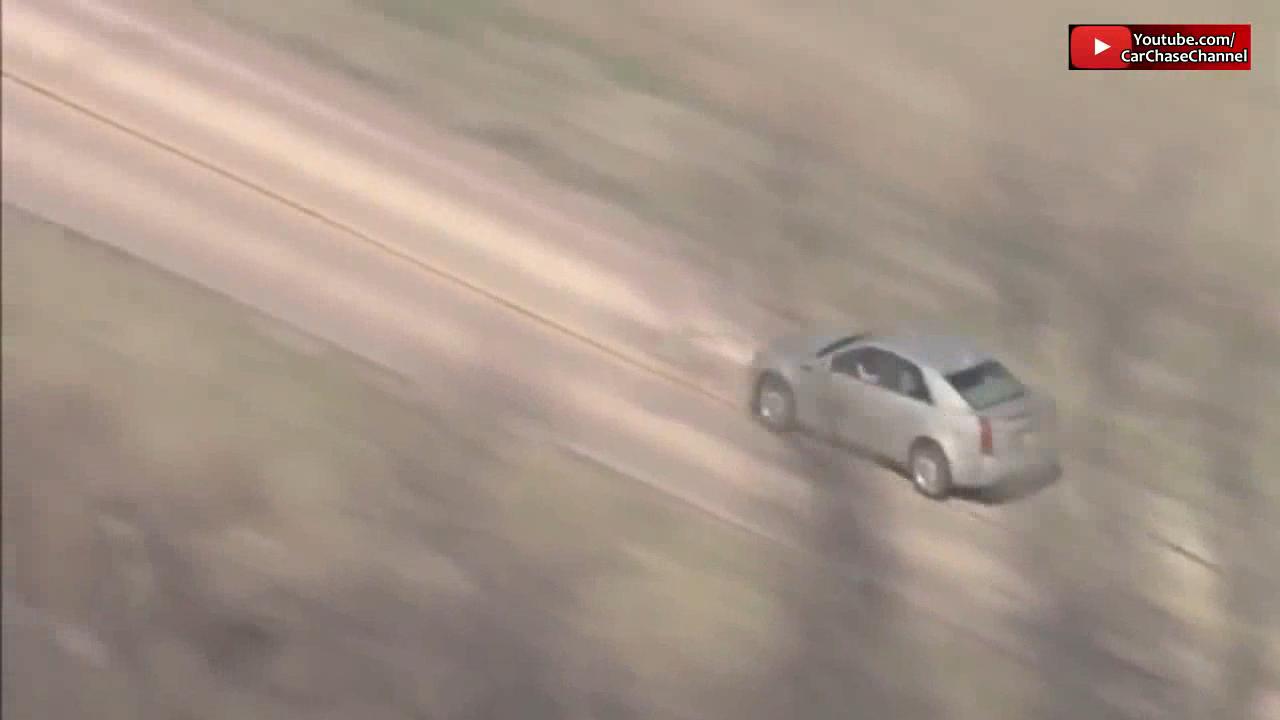}}};
        \node (img1) at (img2.west) {\frame{\includegraphics[height=0.5in]{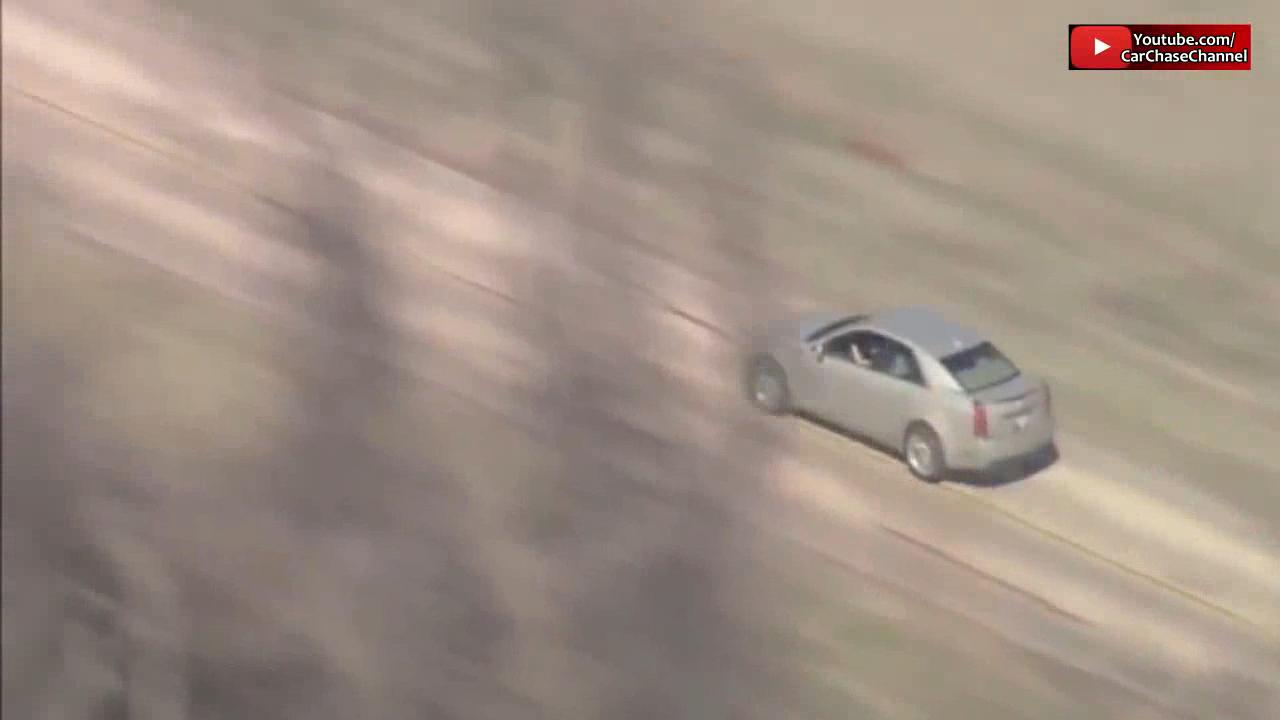}}};
        \node (etc) at (img5.east) {$\mathbf{\ldots}$};
        \node (people) [left =of img1,xshift=0.3in,yshift=-0.4in] {\includegraphics[height=1.2in]{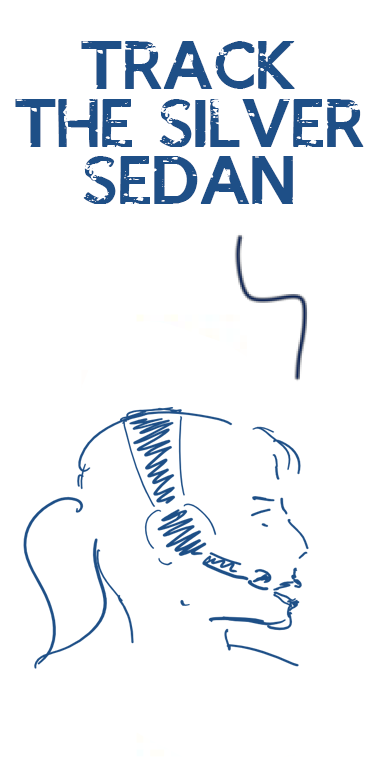}};
        \node (gt5) [below=of img5,yshift=0.3in]{\frame{\includegraphics[height=0.5in]{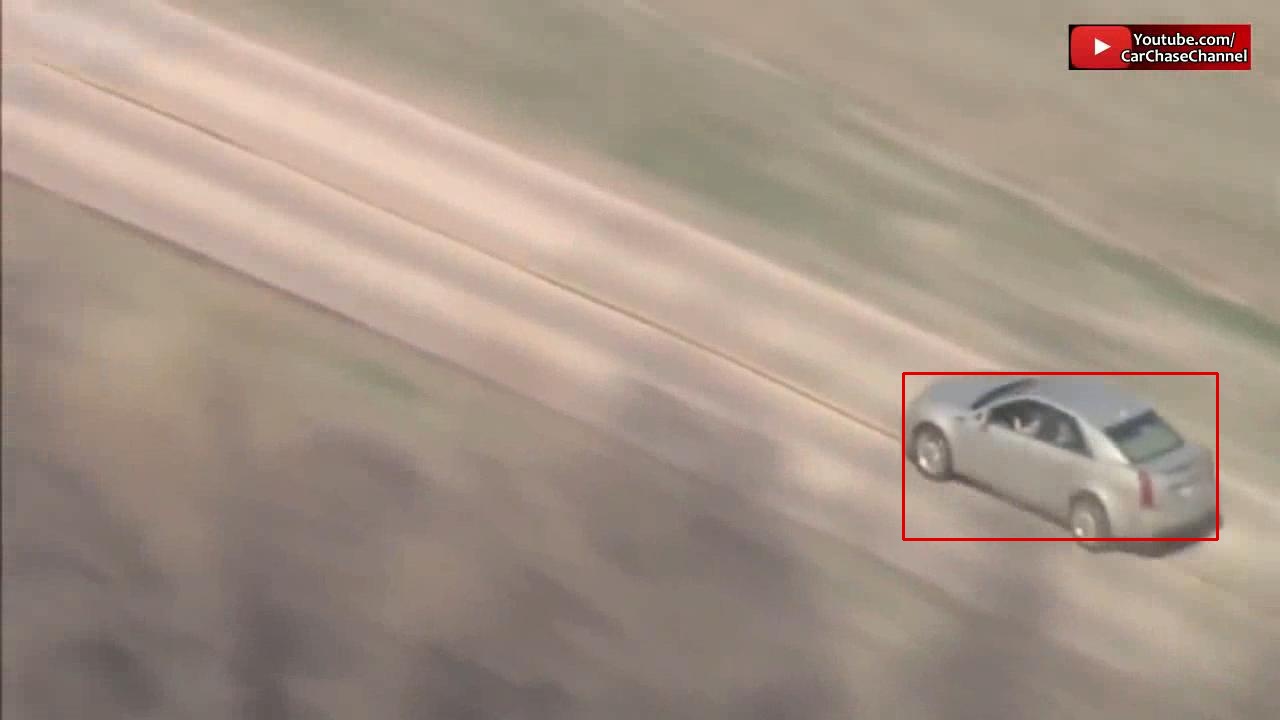}}};
        \node (gt4) at (gt5.west) {\frame{\includegraphics[height=0.5in]{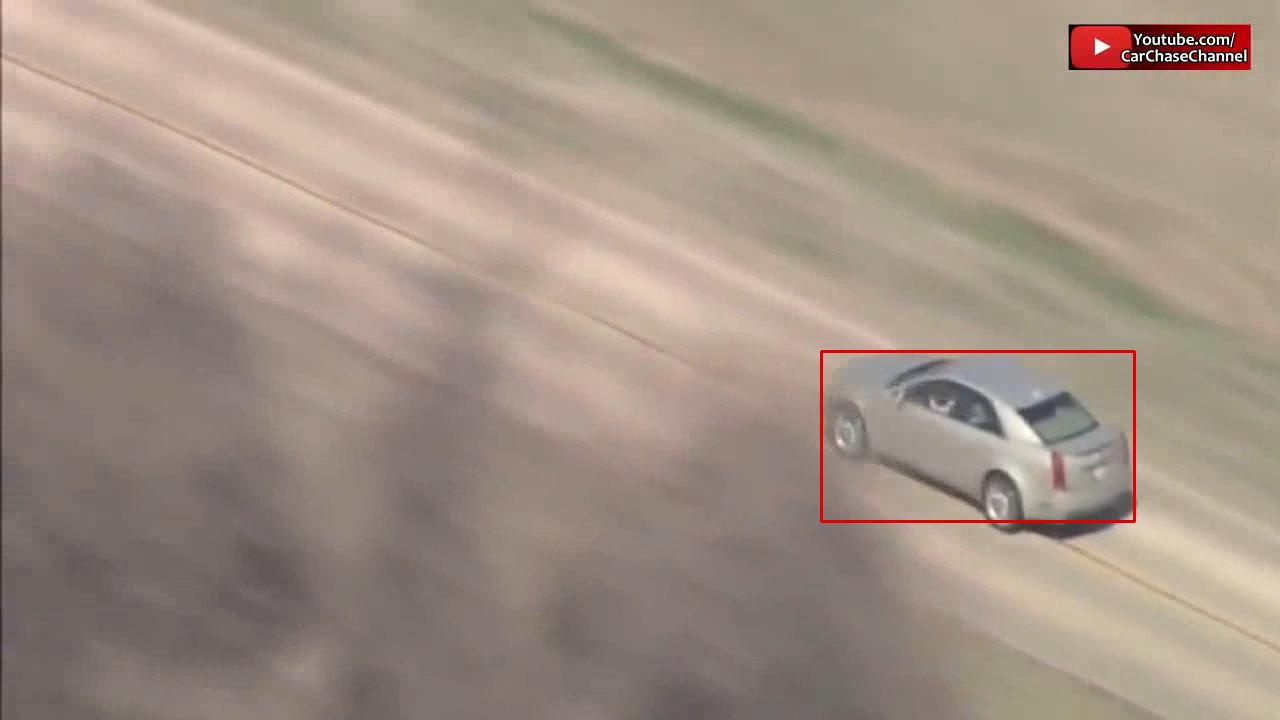}}};
        \node (gt3) at (gt4.west) {\frame{\includegraphics[height=0.5in]{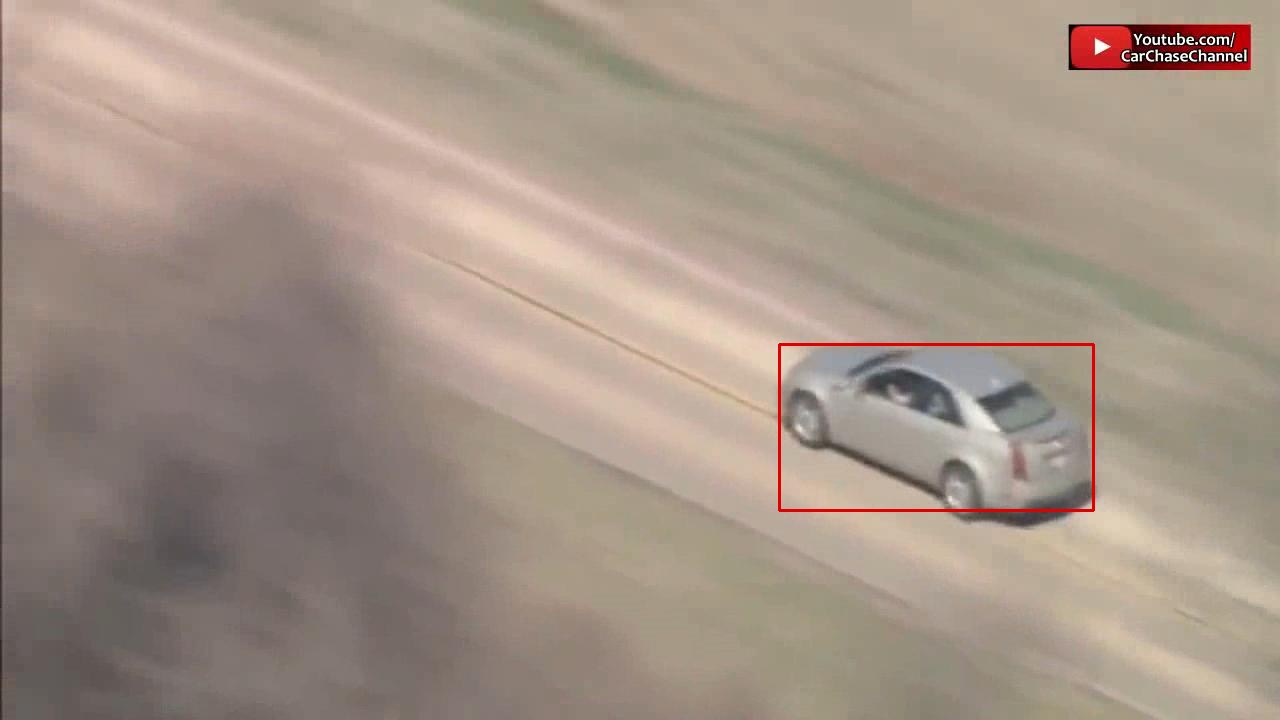}}};
        \node (gt2) at (gt3.west) {\frame{\includegraphics[height=0.5in]{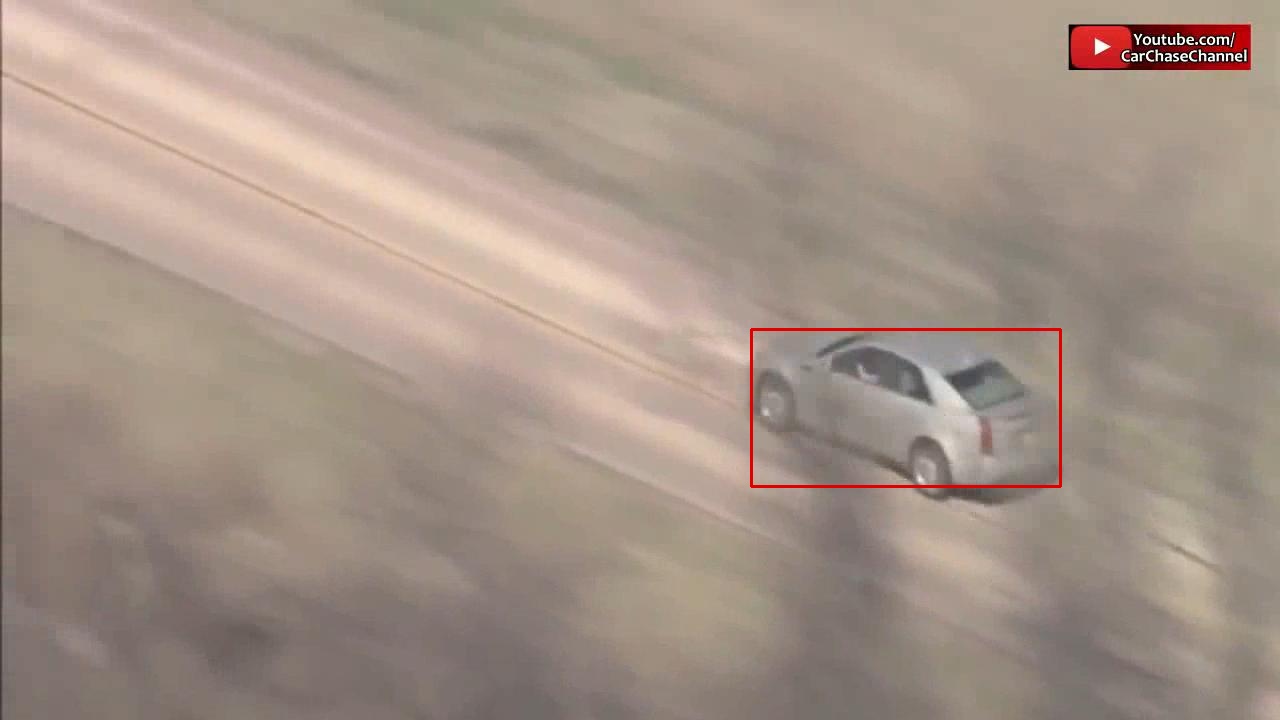}}};
        \node (gt1) at (gt2.west) {\frame{\includegraphics[height=0.5in]{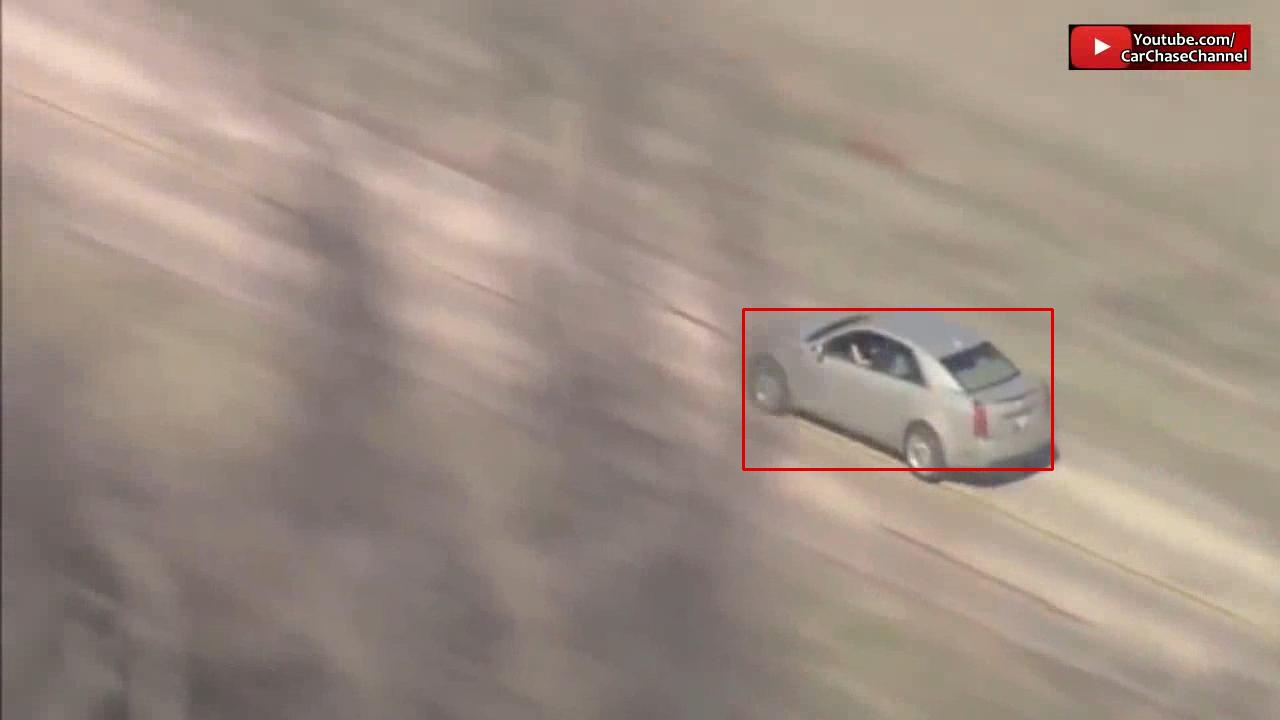}}};
        \node (etc-2) at (gt5.east) {$\mathbf{\ldots}$};
        \coordinate[below=of img1,xshift=0.15in, yshift=-0.05in] (gt1-bbox);
        \coordinate[below=of img2,xshift=0.20in, yshift=-0.06in] (gt2-bbox);
        \coordinate[below=of img3,xshift=0.22in, yshift=-0.07in] (gt3-bbox);
        \coordinate[below=of img4,xshift=0.24in, yshift=-0.08in] (gt4-bbox);
        \coordinate[below=of img5,xshift=0.25in, yshift=-0.09in] (gt5-bbox);
        \draw [arrow,red]  (gt1-bbox) -- (gt2-bbox);
        \draw [arrow,red]  (gt2-bbox) -- (gt3-bbox);
        \draw [arrow,red]  (gt3-bbox) -- (gt4-bbox);
        \draw [arrow,red]  (gt4-bbox) -- (gt5-bbox);
      \end{tikzpicture}
    }
    \caption{The goal is to perform tracking by natural language
      specifications given by a human. For example, someone specifies
      \textit{track the silver sedan running on the highway} and our goal is
      to predict a sequence of bounding boxes on the input video. We also take
      advantage of the natural language to better handle the cases of
      occlusion and rapid motion of the target throughout the tracking process.}
    \label{fig-goal}
  \end{center}
  \vspace{-0.3 in}
\end{figure}

While it may be acceptable, for the sake of benchmark evaluation, to condition
a tracker on an initial bounding box, it limits the practical applicability of
the tracker.  In particular, it assumes forensic-style scenarios where a human
operator sitting at a workstation with a high-definition monitor has the time
and attention to specify an exact bounding box for each object of interest.
Yet, as demand for computer vision evolves, so do requirements and assumptions
for how vision algorithms fit within the intended applications. For example, a
drone operator with a tablet would be unable to draw a precise bounding box on
an object the drone should circle around, e.g., a vehicle speeding down a
highway.  In such a scenario, as is shown in Fig.~\ref{fig-goal}, it is more
acceptable for the operator to speak into the head-microphone the command:
``track the silver sedan.''

Combining natural language understanding and object tracking potentially yields
multiple benefits.  Firstly, it has the ability to dispense with the
requirement of a bounding-box initialization.  Secondly, it combines the
complementary nature of the two modalities, language and visual appearance,
both of which promise the longer-term stability of a tracker.

In spite of the apparent benefits, the role of natural language in tracking has
been relatively unexplored~\cite{Li}.  One of the main challenges is the
derivation of a formulation where a target's location is conditioned both on
its natural language description and its appearance history in the previous
frames.  Another challenge is to guarantee the computational efficiency in
spite of the added complexity. 

Contributions of this paper include:
\begin{itemize}[itemsep=-4pt,topsep=1pt,itemsep=0pt]
  \item
    A novel tracking-by-detection formulation that is conditioned on a natural
    language description of the target to be tracked; in particular, we are the
    first to derive a tracker that employs an RNN to estimate the target's
    location from regions proposed by a deep network, conditioned on a language
    network.  If desired, our formulation can be conditioned on both the NL
    description {\em and} a bounding box, thus providing added flexibility in
    how our tracker can be used in practice.
  \item
    Our tracker nearly doubles the performance of the best prior attempt to
    condition purely on natural language~\cite{Li}, and thus brings us closer
    to making NL tracking practical. It compares favorably to the
    state of the art trackers that employ bounding boxes for initialization in
    cases when NL description is precise.  We show that, for detection based
    trackers, targets can be extended to a wider range of categories with the
    guidance of NL.
  \item
    We implement our tracker following the derivation in Section
    \ref{sec-model} without any additional ``bells-and-whistles,'' such as
    searching across scales, reliance on non-linear dynamics to model target
    motion, or sub-window attention~\cite{song2018vital}. Note that the added
    computational complexity of ~\cite{song2018vital} results in inference
    speed below 1 fps.  Even though we process more information than
    traditional approaches, visual appearance {\em and} NL, our method is
    computationally efficient, yielding a tracker operating at roughly 30 fps
    on a single GPU.
\end{itemize}
\section{Related Works}
Regional convolution networks have seen wide success in recent years.
R-CNN~\cite{Girshick2014}, Fast R-CNN~\cite{Girshick2015}, Faster
R-CNN~\cite{Ren2015} and Masked R-CNN~\cite{He2017} are a family of regional
convolution networks developed for object detection.  The introduction of the
region proposal network (RPN) in the Faster R-CNN framework~\cite{Ren2015}
makes the generation of object proposals efficient and fast compared to the
selective search approach~\cite{Uijlings2013} used in
R-CNN~\cite{Girshick2014}. The RPN in Faster R-CNN is introduced to improve the
speed of proposing regions based on the appearance of each region, while the
object detection module remains the same as in~\cite{Girshick2015}.

The proposal module in our tracker is built upon the RPN, which proposes
regions from convolution features. A Region of Interest (RoI) sampling is
applied to regional features after the RPN to reduce the size of region
proposals and to extract the regional features.

Various models have been proposed to address the problem of object detection
with natural language (NL)
specification~\cite{Boom2016,Johnson2016,Lee2017,Li2017,Vinyals2016}. Our work
employs NL understanding techniques similar to some image captioning approaches
to perform the task of object detection with NL specification.

In recent work, word embeddings~\cite{Pennington2014} are used to represent
words in numerical vector representations. In our work, word embeddings are
used for measuring the similarities between NL specifications and images.
Johnson et al.\ proposed a dense captioning model~\cite{Johnson2016} based on
an RPN like region proposal module for region proposals and used a bilinear
sampler to obtain a list of regional features for each of the best proposals
after the RPN. The regional features are further used as the initial state in
an LSTM model for captioning tasks, which is similar to the image captioning
model in Vinyals and others' work~\cite{Vinyals2016}.

In our work, we use the same approach to obtain region proposals and regional
features to score and sample the detections for tracking.

Tracking by detection models~\cite{Blackman2004,Kalal2012} and
Bayesian filtering based algorithms~\cite{Brookner1998,Kalman1960} have been
widely adopted. Some deep learning based
models~\cite{bertinetto2016fully,danelljan2017eco,ning2017spatially,ning2017spatially,song2018vital,Wang2013}
have been introduced in recent years, and are argued to perform better when
handling occlusion and appearance change. SiamFC~\cite{bertinetto2016fully}
conducts a local search for regions with a similar regional visual features
obtained by a CNN in every frame. ECO~\cite{danelljan2017eco} applies
convolutional filters on convolution feature maps to obtain satisfactory
performance on multiple tracking datasets. ECO still suffers from efficiency
issues~\cite{huang2017learning}, though its efficiency is improved from the
original convolution filter based tracker, C-COT~\cite{danelljan2016beyond}.
These trackers maintain appearance and motion models explicitly by maintaining
the visual features over time.
Other approaches to the visual object tracking problem employ recurrent neural
networks (RNNs).  ROLO~\cite{ning2017spatially} attempts to model the temporal
relationships between detections across continuous frames. The RNN performs a
regression task given detections as input.

In contrast, NL based tracking allows the tracker to track a wider range of
objects. Li et al.~introduced three models that exploit NL descriptions of the
target~\cite{Li}, which jointly consider the regional appearance features and
NL features. They proposed a way to initialize a tracker without a bounding
box, although it is 50\% less accurate compared to ground truth bounding box initialized tracker.

Our model performs the same tracking by NL task in a different way by jointly perform the
tracking and detection at the same time. Comparing to \cite{Li}, instead of
only using the NL description at the first frame to initialize the tracker, we
utilize the NL description during the entire tracking process.

\section{The Proposed Model}~\label{sec-model}
\begin{figure}[t]
  \begin{center}
    \resizebox{\columnwidth}{!}{
      \begin{tikzpicture}[node distance=0.2in]

        \node (in-frame-1) [startstop] {$I_{t-1}, Q$};
        \node (in-frame-2) [startstop, right=of in-frame-1,xshift=1 in] {$I_{t}, Q$};
        % \node (in-query) [startstop, left=of in-frame-1,xshift=-0.5in] {$Q$};

        \node (in-png-1) [below=of in-frame-1] {\includegraphics[width=1.5in]{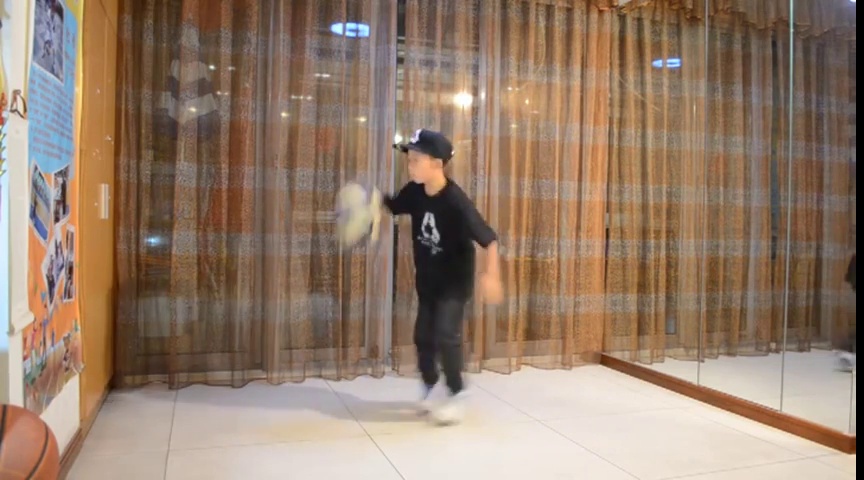}};
        \node (in-png-2) [below=of in-frame-2] {\includegraphics[width=1.5in]{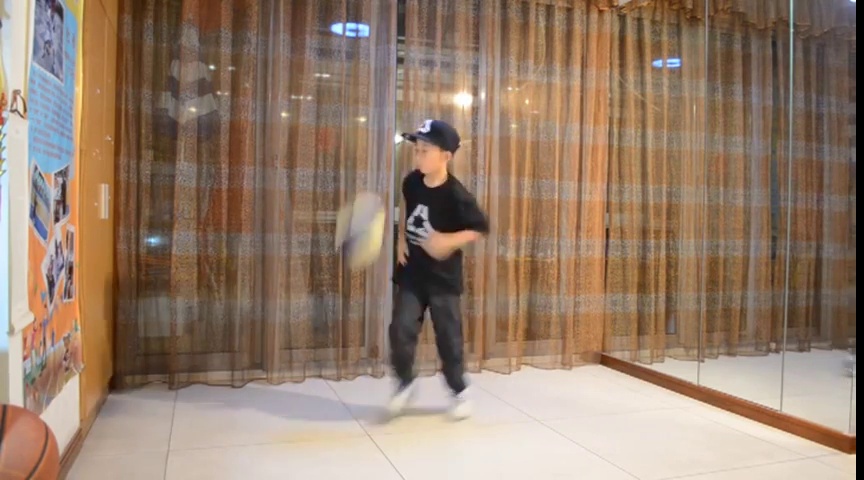}};
        % \node (in-q) [below=of in-query,yshift=-0.2in] {The yellow and black ball.};

        \node (frame-1-lang) [io, above=of in-frame-1] {$\mathbf{X}_{t-1}^\text{det}$, $\mathbf{RF}_{t-1}^\text{det}$};
        \node (frame-2-lang) [io, above=of in-frame-2] {$\mathbf{X}_{t}^\text{det}$, $\mathbf{RF}_{t}^\text{det}$};
        \draw [arrow] (in-frame-1) -- (frame-1-lang) node[midway,left]{detection} node[midway, right]{phase};
        \draw [arrow] (in-frame-2) -- (frame-2-lang) node[midway,left]{detection} node[midway, right]{phase};

        \node (frame-1-lstm) [process, above=of frame-1-lang] {Tracking Module};
        \node (frame-2-lstm) [process, above=of frame-2-lang] {Tracking Module};
        \draw [arrow] (frame-1-lang) -- (frame-1-lstm);
        \draw [arrow] (frame-2-lang) -- (frame-2-lstm);
        \draw [arrow] (frame-1-lstm) -- (frame-2-lstm) node[midway, above] {hidden} node[midway, below] {state};

        \node (frame-1-track) [startstop, above=of frame-1-lstm] {$\hat{X}_{t-1}$};
        \node (frame-2-track) [startstop, above=of frame-2-lstm] {$\hat{X}_{t}$};
        \draw [arrow] (frame-1-lstm) -- (frame-1-track)node[midway,left] {output};
        \draw [arrow] (frame-2-lstm) -- (frame-2-track)node[midway,left] {output};
        \draw [arrow] (frame-1-lstm) -- (frame-2-lang) node [midway,below,yshift=-0.1 in] {Relaxed Gating};

        \node (out-png-1) [above=of frame-1-track] {\includegraphics[width=1.5in]{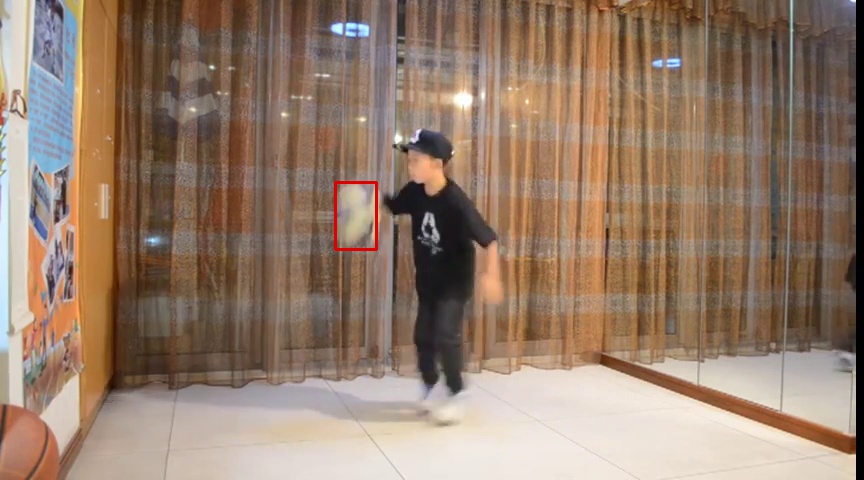}};
        \node (out-png-2) [above=of frame-2-track] {\includegraphics[width=1.5in]{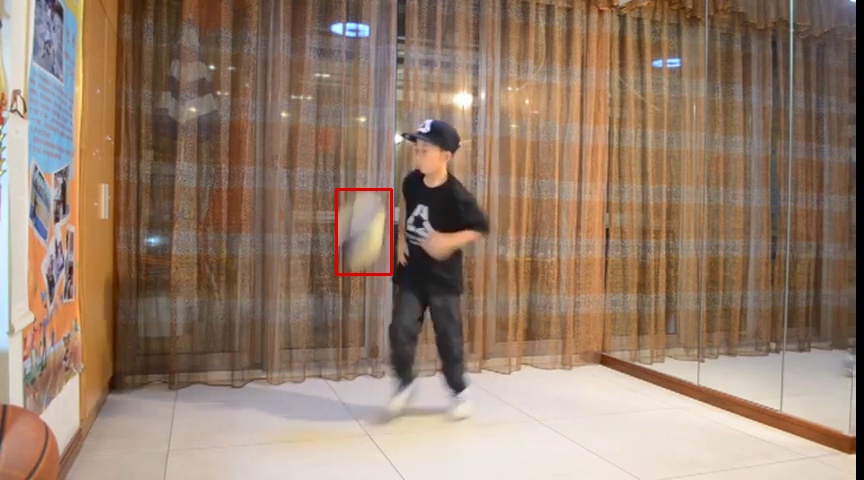}};
      \end{tikzpicture}
    }
    \caption{The proposed model. Red rounded rectangles are inputs and
      outputs of the model. The detection phase (detailed in Fig.~\ref{fig-detection-phase}) of the model takes frames and NL descriptions as inputs
      and returns a set of candidate bounding boxes and their corresponding region features. The tracking module performs tracking by detection.
      Predictions are then utilized to assist the detection phase with
      relaxed gating.}
    \label{fig-architectures}
    \vspace{-0.2 in}
  \end{center}
\end{figure}

Our proposed tracker is composed of two phases. The first phase is the detection phase which contains two modules. The first module is a deep
convolutional neural network that proposes candidate detection regions, and the
second module is an RNN that measures the similarities between visual
representations of proposed regions from the first module and the input
language embeddings~\cite{Pennington2014}. After the detection phase, regions are ranked by the
similarities and passed into the tracking module, which is another RNN that
performs tracker update from previous frame based on detection candidates.
The overall architecture of our model is shown in Fig.~\ref{fig-architectures},
expressed as a Bayesian network, while the detection phase is decoupled from the
tracker and is detailed in Fig.~\ref{fig-detection-phase}.

\begin{table}
  \begin{center}
    \resizebox{\columnwidth}{!}{
      \begin{tabular}{rl}
        \hline
        \hline
        \multicolumn{2}{c}{\textbf{Deterministic Variables}}\\
        \hline
        \hline
        $t$ & frame index \\
        $I_t$ & the $t$-th frame image from a video\\
        $Q$ & the given natural language query for a video\\
        $X_t^*$ & target (ground truth) bounding box.\\
        \hline
        \hline
        \multicolumn{2}{c}{\textbf{Random Variables}}\\
        \hline
        \hline
        $F_t$ & visual feature map computed by CNN \\
        $\mathbf{X}_t=\{X_{t,i}\}$ & set of bounding boxes proposed by RPN \\
        $\mathbf{RF}_t=\{RF_{t,i}\}$ & set of region features for bounding boxes in $\mathbf{X}_t$\\
        $\mathbf{S}_t=\{S_{t,i}\}$ & set of similarities between region features and $Q$\\
        $\mathbf{X}_t^\text{det}$ & set of bounding boxes chosen by detection module\\
        $\mathbf{RF}_t^\text{det}$ & set of region features for bounding boxes in $\mathbf{X}_t^\text{det}$\\
        $\hat{X_t}$ & bounding box predicted by the tracker \\
        \hline
      \end{tabular}
    }
  \end{center}
  \caption{Notations of deterministic and random variables.}
  \label{tbl-variables}
  \vspace{-0.1 in}
\end{table}

In Section~\ref{sec-detections}, we
formulate the detection problem and compute the posterior probability of
a detection given the target's NL description. In Section~\ref{sec-tracking}, we present
our tracking module. Finally, we present implementation details in
Section~\ref{sec-details}.  Notations that we use in deriving our tracker are
summarized in Table~\ref{tbl-variables}.

\subsection{Detection}~\label{sec-detections}
Let \(I_1, \ldots, I_T\) denote the sequence of video frames, where
\(T\) is the total number of frames. Let \(Q\) be the NL description of the target.
Let \(\mathbf{X}_t = \{X_{t,i}\}^n_{i=1}\) be the set of candidate bounding
boxes.

\begin{algorithm}[htbp]
  \While{Training RPN}{
    \(\mathbf{X}_t \leftarrow F_\text{RPN} (I_t) \)\;
    \(L_\text{RPN} \leftarrow \text{RPN Loss}(\mathbf{X}_t, X_t^*)\)\;
    Backprop(\(L_\text{RPN}\))\;
  }
  Fix(\(F_\text{RPN}\))\;
  \While{Training Language Network}{
    \(\mathbf{X}_t \leftarrow F_\text{RPN} (I_t) \)\;
    \(\mathbf{X}_t^\text{det}, \mathbf{S_t} \leftarrow F_\text{lang} (\mathbf{X}_t) \)\;
    \(L_\text{lang} \leftarrow \text{LangNet Loss}(\mathbf{X}^\text{det}_t, X_t^*, Q, \mathbf{S}_t)\)\;
    Backprop(\(L_\text{lang}\))\;
  }
  Fix(Language Network)\;
  \While{Training Tracker}{
    \For{\(t\in \{1,\cdots,T\}\)}{
      \(\mathbf{X}_t^\text{det}, \mathbf{RF}_t^\text{det} \leftarrow \text{Top-K}(\mathbf{X}_t)\)\;
      \(\hat{X_t} \leftarrow F_\text{tracker} (\mathbf{X}_t^\text{det},\mathbf{RF}_t^\text{det}) \)\;
    }
    \(L_\text{tracker} \leftarrow \sum_{t\in \{1,2,\cdots,T\}}\text{Tracker Loss}(\hat{X}_t, X_t^*)\)\;
    Backprop(\(L_\text{tracker}\))\;
  }
  \caption{Curriculum training for the tracker.}
  \label{alg-training}
\end{algorithm}

\begin{algorithm}[htbp]
  \If{Initial Bounding Box is Known}{
    \(\hat{X_1} \leftarrow X_1^*\)\;
  }
  \Else{
    \(\mathbf{X}_1 \leftarrow F_\text{RPN}(I_1)\)\;
    \(\mathbf{RF}_1 \leftarrow \text{ROI Pooling}(\mathbf{X}_1, F_1)\)\;
    \(\mathbf{S}_1 \leftarrow F_\text{lang} (\mathbf{RF}_1, Q)\)\;
    ReRank(\(\mathbf{X}_1, \mathbf{S_1}\))\;
    \(\mathbf{X}_1^\text{det}, \mathbf{RF}_1^\text{det} \leftarrow \text{Top-K}(\mathbf{X}_1)\)\;
    \(\hat{X_1} \leftarrow F_\text{tracker}(\mathbf{X}_1^\text{det}, \mathbf{RF}_1^\text{det})\)\;
  }
  \For{\(I_t \in \{I_2, I_2,\cdots, I_T\}\)}{
    \(\mathbf{X}_t \leftarrow F_\text{RPN}(I_t)\)\;
    \(\mathbf{RF}_t \leftarrow \text{ROI Pooling}(\mathbf{X}_t, F_t)\)\;
    \(\mathbf{S}_t \leftarrow F_\text{lang} (\mathbf{RF}_t, Q)\)\;
    \(\mathbf{S}_t \leftarrow \text{Relaxed Gating}(\hat{X}_{t-1}) \cdot \mathbf{S}_t\)\;
    ReRank(\(\mathbf{X}_t, \mathbf{S_t}\))\;
    \(\mathbf{X}_t^\text{det}, \mathbf{RF}_t^\text{det} \leftarrow \text{Top-K}(\mathbf{X}_t)\)\;
    \(\hat{X_t} \leftarrow F_\text{tracker}(\mathbf{X}_t^\text{det}, \mathbf{RF}_t^\text{det})\)\;
  }
  \caption{Inference using the proposed tracker.}
  \label{alg-inference}
  \KwResult{\(\hat{X_1}, \cdots, \hat{X_T}\)}
\end{algorithm}

Given frame \(I_t\), a deep convolutional neural network is utilized to extract
the visual feature map, which we denote as \(F_t\in \mathbb{R}^{H\times W\times
D}\), where \(H, W, \text{and } D\) are height, width, and depth of the
visual feature map.

A region of interest (RoI) pooling on \(F_t\) is then adopted to obtain
regional convolution features of a uniform size \(A \times A \times D\) for
all bounding boxes in \(\mathbf{X}_t\).  We denote the set of regional
convolution features as \(\mathbf{RF}_{t}=\{RF_{t,i}\}^n_{i=1}\), where
\(RF_{t,i} \in \mathbb{R}^{A\times A\times D}\).

We adopt a Region Proposal Network (RPN) that performs class agnostic object
detection on $\mathbf{RF}_t$. We consider this RPN as an estimation of a
conditional probability $\Pr[X_{t,i}|I_t]$.

Similarly to the RPN, we design a language network that estimates the similarity
between \(Q\) and each \(X_{t,i}\), which can be regarded as the conditional
probability \(\Pr[X_{t,i}|Q]\) for each bounding box.

Our language network is an LSTM network that takes word embeddings from
GloVe~\cite{Pennington2014} as inputs. The language network learns a sentence
embedding of size $300$, which is also the size of the embedding of GloVe. The
language module then produces a similarity score \(S_{t,i}\) between the
sentence embedding from $Q$ and \(RF_{t,i}\).

We train our language network as a regression task with a log loss. The target
similarity \(\hat{S}_{t,i}\) is the intersection-over-union (IoU) between
\(X_{t,i}\) and the ground truth bounding box \(X^*_t\).  That is

\begin{equation}
  L_\text{lang}(t) = -\sum\left(\hat{\mathbf{S}}_{t}\log (\mathbf{S}_{t}) +
  (1-\hat{\mathbf{S}}_{t})\log(1-\mathbf{S}_{t})\right),
\end{equation}
where
\begin{equation*}
  \hat{\mathbf{S}}_{t} = \text{IoU}(\mathbf{X}_{t}, X^*_t).
\end{equation*}

To make computation more efficient, we only compute the language similarity for
bounding boxes of \(\mathbf{X}_t\) that are not among the top \(N=5\) during
inference.

After estimating the joint conditional probability \(\Pr[X_{t,i}|Q,I_t]\) for
all candidates \(X_{t,i} \in \mathbf{X}_t\) proposed by the RPN following the
Bayes theory, we rank the bounding boxes by these new scores.  We choose a set
of the top \(N=5\) bounding boxes and denote this set as
\(\mathbf{X}^\text{det}_t\).  The detection set for each frame
\(\mathbf{X}^\text{det}_1, \ldots, \mathbf{X}^\text{det}_T\), and their
corresponding sets of regional convolution features
\(\mathbf{RF}_1^\text{det},\ldots,\mathbf{RF}^\text{det}_T\) are further
pipelined into the tracking module to perform tracking sequentially.

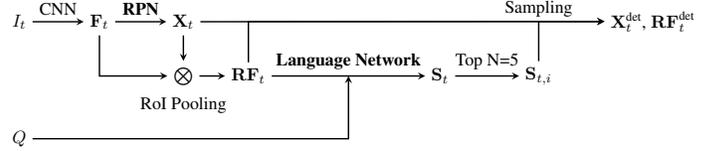
\begin{figure}[t]
  \begin{center}
    \resizebox{\columnwidth}{!}{
      \begin{tikzpicture}[node distance=0.2in]
        \node (in-frame) [] {$I_{t}$};
        \node (in-query) [below=of in-frame,yshift=-0.5in] {$Q$};
        \node (cnn-output) [right=of in-frame,xshift=0.2in] {$\mathbf{F}_{t}$};
        \node (rpn-output) [right=of cnn-output,xshift=0.2in] {$\mathbf{X}_{t}$};
        \node (place-holder-roi) [below=of rpn-output] {$\bigotimes$};
        \node (roi-pooling) [below=of place-holder-roi,yshift=0.2 in] {RoI Pooling};
        \node (roi-output) [right=of place-holder-roi] {$\mathbf{RF}_t$};
        \coordinate (lang-net) at ([xshift=0.6in]roi-output.east);
        \node (lang-sim) [right=of lang-net,xshift=0.4in] {$\mathbf{S}_t$};
        \node (top-lang-sim) [right=of lang-sim,xshift=0.3in] {${\mathbf{S}}_{t,i}$};
        \node (sample) [above=of top-lang-sim, yshift=0.1in] {Sampling};
        \node (result) [right=of sample,yshift = -0.1in] {$\mathbf{X}_t^\text{det}$, $\mathbf{RF}_t^\text{det}$};

        \draw [arrow] (in-frame) -- (cnn-output) node[midway,above]{CNN};
        \draw [arrow] (cnn-output) -- (rpn-output) node[midway,above]{\bf RPN};
        \draw [arrow] (rpn-output) -- (place-holder-roi);
        \draw [arrow] (cnn-output) |- (place-holder-roi);
        \draw [arrow] (place-holder-roi) -- (roi-output);
        \draw [arrow] (roi-output) -- (lang-sim) node[midway,above]{\bf Language Network};
        \draw [arrow] (in-query) -| (lang-net);
        \draw [arrow] (lang-sim) -- (top-lang-sim) node[midway,above]{Top N=5};
        \draw [arrow] (rpn-output) -- (result);
        \draw [arrow] (roi-output) |- (result);
        \draw [arrow] (top-lang-sim) |- (result);
      \end{tikzpicture}
    }
     \caption{The architecture of the detection phase, which consumes inputs
       (frame $I_t$ and natural language description $Q$) and produces bounding
       boxes \(\mathbf{X}_t^\text{det}\) and their corresponding regional
       convolution features \(\mathbf{RF}_t^\text{det}\) chosen by the RPN and
       the language network. The detection phase is the first phase in our
       model, preceding the tracking module in Fig.~\ref{fig-architectures}.}
    \label{fig-detection-phase}
    \vspace{-0.2 in}
  \end{center}
\end{figure}

The
architecture of the detection phase is summarized and shown in
Fig.~\ref{fig-detection-phase}.

\subsection{Tracking}\label{sec-tracking}
A recurrent neural network (RNN) is well-suited for object tracking due to its
ability to maintain model states.  In particular, we use a long short-term
memory (LSTM) architecture to maintain the motion, position and appearance
history of the target in its hidden-state representation. Detections
\(\mathbf{X}_t^\text{det}\) and their corresponding regional convolution
features \(\mathbf{RF}_t^\text{det}\) are recurrently fed into the tracking module as
inputs. We apply a fully connected layer with four neurons on LSTM outputs to
obtain the track for time \(1,\ldots,T\).  We use \(\hat{X_t}\) to denote the
tracker prediction at frame \(I_t\).  The tracking module is trained as a
regression task with smoothed L1 loss~\cite{Ren2015} for each time step.

\textbf{Relaxed Gating:} During inference of the detection phase of our
tracker, we apply a relaxed gating to \(\mathbf{S}_t\) within the detection
module.

We use the difference between \(\hat{X}_{t-1}\) and \(\hat{X}_{t-2}\) to
measure the velocity $v$ for our target. We measure the distance $\mathbf{d}_t$
from \(\hat{X}_{t-1}\) and boxes from detection module
\(\mathbf{X}_t^\text{det}\).  Similarities from the language module is then
updated to ``reflect" gating on the motion.

\begin{equation}
  \mathbf{S}_t^\text{relaxed gating} \leftarrow
  \mathbf{S}_t \cdot v / \mathbf{d}_t.
\end{equation}

Relaxed gating is applied to bounding box proposals during evaluation and
training of the tracking module.  The goal of relaxed gating is to suppress
signals from objects that are related to the NL description but are not in the
target's trajectory.

\subsection{Implementation Details}~\label{sec-details}
We train and test our detection and tracking modules with original resolution
images and NL annotations. We crop images so that the longest side of the image
is 1333 pixels~\cite{Detectron2018}. For anchors, we use 5 sizes at
\([32^2,64^2,128^2,256^2,512^2]\), which is slightly larger than the original
Faster-RCNN~\cite{Ren2015}, to accommodate the VisualGenome~\cite{Krishna2016}
dataset. Three different aspect ratios of 1:1, 1:2, and 2:1 are used.

We adopt a three stage training algorithm to learn the parameters in our RPN,
language network, and tracking module. The training algorithm is shown in
Alg.~\ref{alg-training}.

\begin{figure*}[t]
  \centering
  \begin{subfigure}[b]{0.325\textwidth}
    \includegraphics[width=\textwidth]{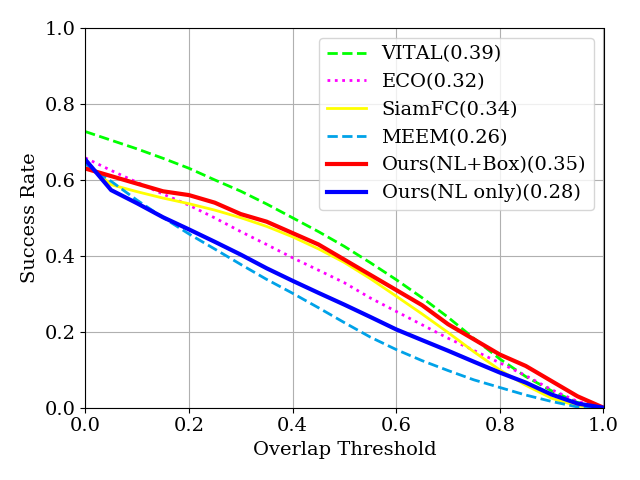}
    \caption{Success Plots on LaSOT}
    \label{fig-lasot-success-performance}
  \end{subfigure}
  \begin{subfigure}[b]{0.325\textwidth}
    \includegraphics[width=\textwidth]{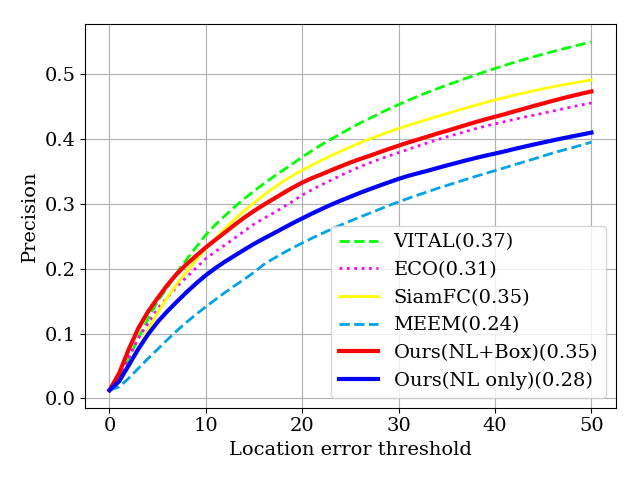}
    \caption{Precision Plots on LaSOT}
    \label{fig-lasot-precision-performance}
  \end{subfigure}
  \begin{subfigure}[b]{0.325\textwidth}
    \includegraphics[width=\textwidth]{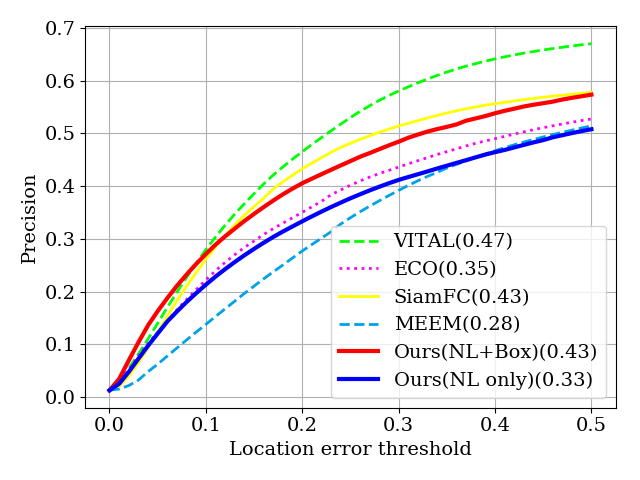}
    \caption{Norm. Precision Plots on LaSOT}
    \label{fig-lasot-normed-precision-performance}
  \end{subfigure}
  \vskip\baselineskip
  \begin{subfigure}[b]{0.325\textwidth}
    \includegraphics[width=\textwidth]{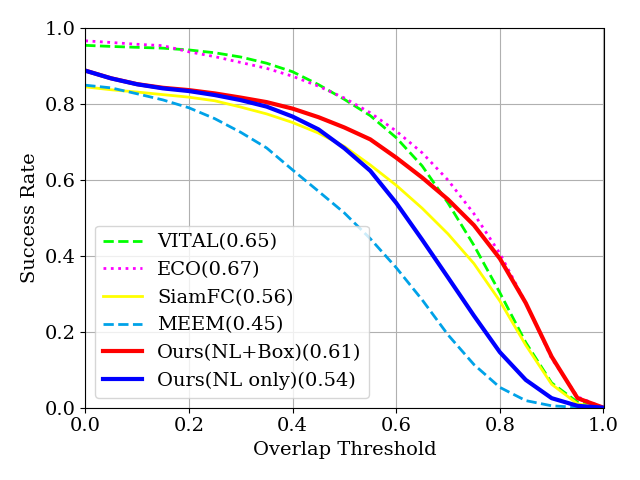}
    \caption{Success Plots on OTB}
    \label{fig-otb-success-performance}
  \end{subfigure}
  \begin{subfigure}[b]{0.325\textwidth}
    \includegraphics[width=\textwidth]{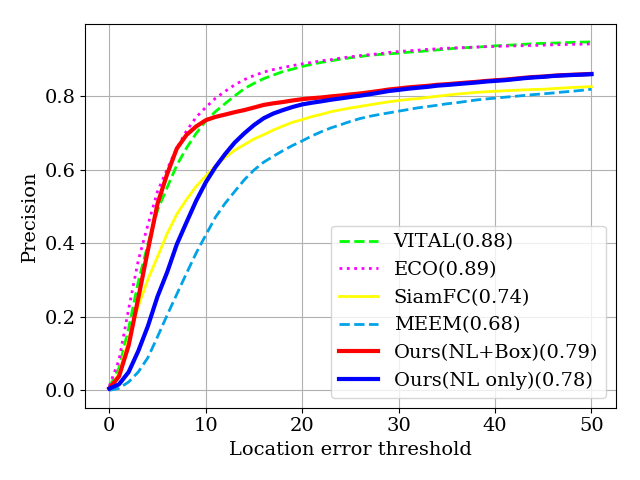}
    \caption{Precision Plots on OTB}
    \label{fig-otb-precision-performance}
  \end{subfigure}
  \begin{subfigure}[b]{0.325\textwidth}
    \includegraphics[width=\textwidth]{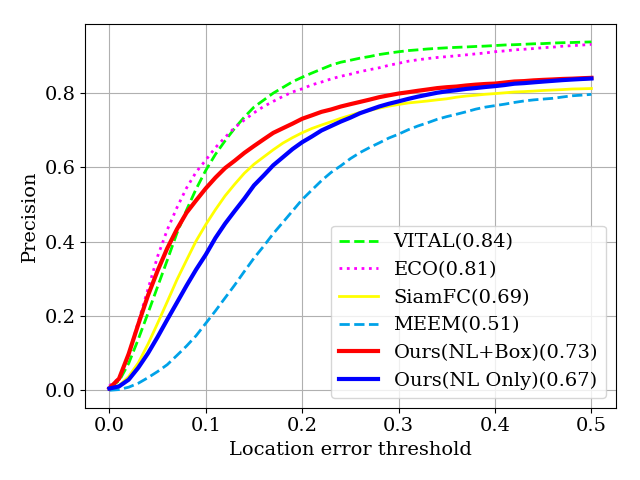}
    \caption{Normalized Precision Plots on OTB}
    \label{fig-otb-normed-precision-performance}
  \end{subfigure}
  \caption{Comparison of our approach with state-of-the-art traditional
    trackers: one pass evaluation success rate at different intersection over
    union thresholds and mean average precision / mean average normalized
    precision at different location error thresholds on testing videos in LaSOT
    and OTB-99-lang.}
    \vspace{-0.2 in}
  \label{fig-performance}
\end{figure*}

In the first stage, we train the RPN as described in Faster
RCNN~\cite{Ren2015}.  We use the same loss used in training RPN in Faster
RCNN.  After the RPN is trained, we freeze parameters in the RPN and train the
second stage, our language network, using the loss described in
Section~\ref{sec-detections}. At this point, our model is trained to
propose candidate regions that are related to the NL description, i.e.
\(\mathbf{X}_t^\text{det}\).  The last stage entails training the tracking module. As
we freeze parameters in both the RPN and the language network, we apply relaxed
gating to the RPN and language detection as described in
Section~\ref{sec-tracking}. Region convolution features together with bounding
box predictions are fed into our tracking module. The tracking module, described in
Section~\ref{sec-tracking}, is finally trained as a regression task on a
smoothed L1 loss~\cite{Ren2015}.

The algorithm for inference is shown in Alg.~\ref{alg-inference}. If the
initial bounding box is not given, the RPN, the language network and the tracking module work
the same way as in other frames, except for the fact that no relaxed gating is
applied to initialize the tracker.

We use a learning rate starting at 0.01 and AdaGrad~\cite{duchi2011adaptive} for
training the model. The proposed tracker is trained with a
TensorFlow~\cite{abadi2016tensorflow} 1.13 implementation.

\section{Experiments}~\label{sec-experiments}
The goal of our experiments is to demonstrate the advantage of our proposed
approach in utilizing NL descriptions either independently or jointly with
appearance cues in visual tracking. We first compare our model against the best
published NL-based tracker~\cite{Li} in
Section~\ref{sec-lang-tracker-performance}. In Section~\ref{sec-performance},
we conclude that with the help of language cues a tracker can reacquire targets
and boost performance, with comparisons between state of the art trackers that
employ bounding boxes for initialization. Finally, in
Section~\ref{sec-ablation-studies}, we conduct ablation studies on our tracker.

\begin{figure*}[t]
  \centering
  \begin{subfigure}[b]{0.325\textwidth}
    \includegraphics[width=\textwidth]{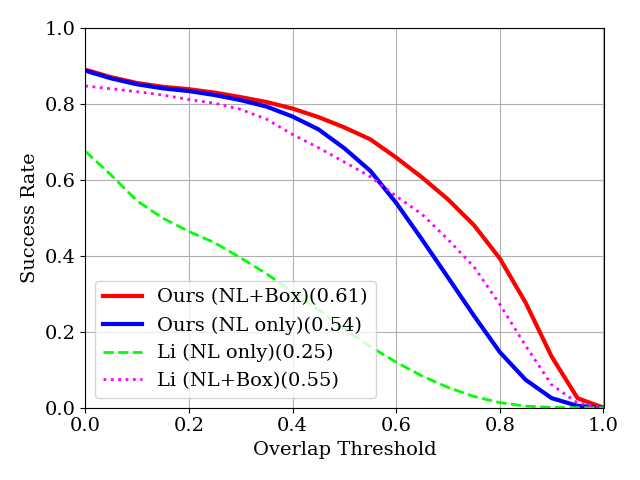}
    \caption{Success Plots on OTB}
    \label{fig-nl-success-performance}
  \end{subfigure}
  \begin{subfigure}[b]{0.325\textwidth}
    \includegraphics[width=\textwidth]{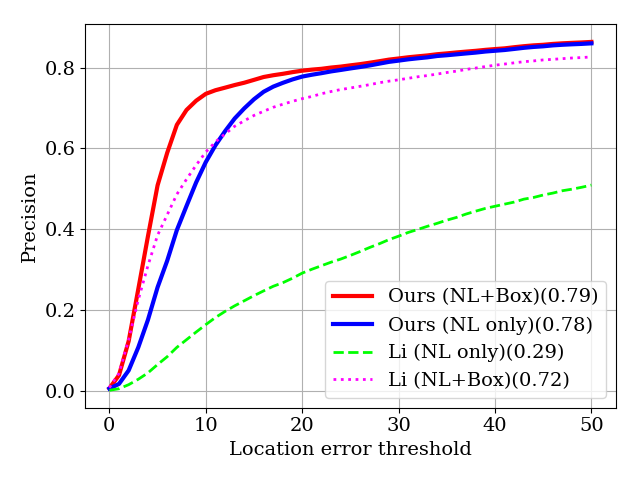}
    \caption{Precision Plots on OTB}
    \label{fig-nl-precision-performance}
  \end{subfigure}
  \begin{subfigure}[b]{0.325\textwidth}
    \includegraphics[width=\textwidth]{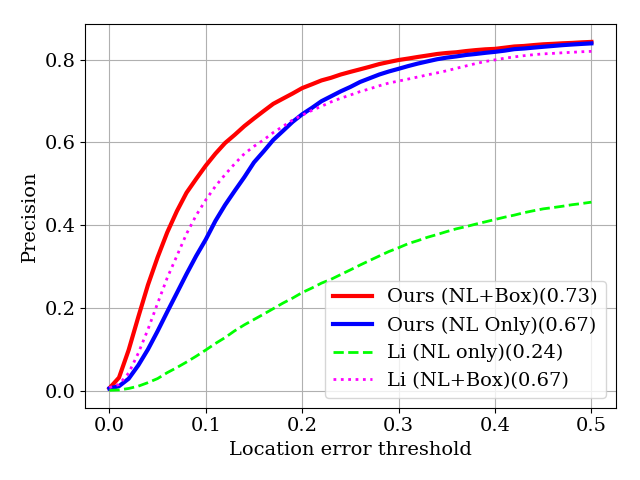}
    \caption{Normalized Precision Plots on OTB}
    \label{fig-nl-normed-precision-performance}
  \end{subfigure}
  \vspace{-0.1 in}
  \caption{Our method vs.~prior best NL tracker Li~\cite{Li} on OTB-99-language
    dataset using the OPE protocol.}
  \label{fig-nl-tracker}
\end{figure*}

\begin{figure*}[!t]
  \centering
  \begin{subfigure}[b]{0.325\textwidth}
    \includegraphics[width=\textwidth]{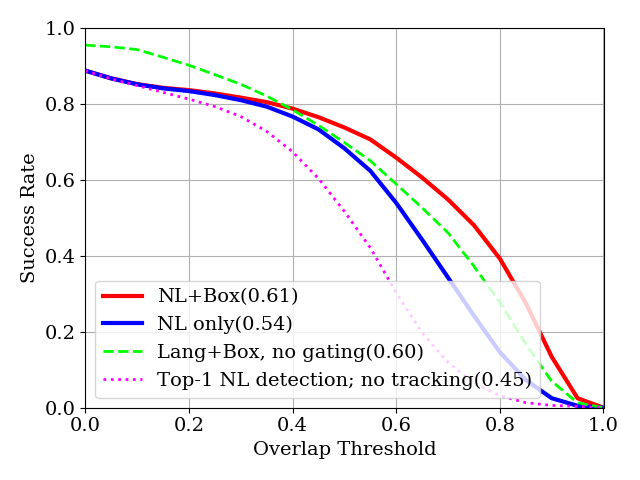}
    \caption{Success Plots on OTB}
    \label{fig-otb-ablation-success-performance}
  \end{subfigure}
  \begin{subfigure}[b]{0.325\textwidth}
    \includegraphics[width=\textwidth]{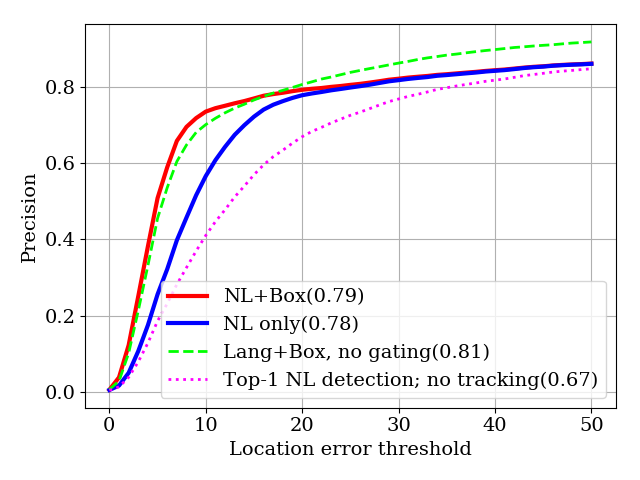}
    \caption{Precision Plots on OTB}
    \label{fig-otb-ablation-precision-performance}
  \end{subfigure}
  \begin{subfigure}[b]{0.325\textwidth}
    \includegraphics[width=\textwidth]{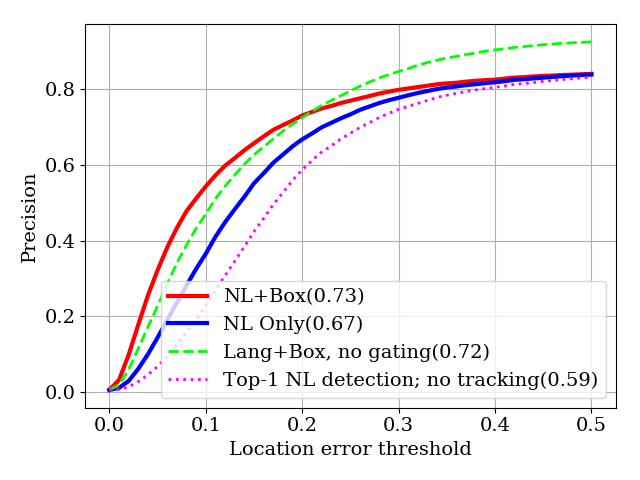}
    \caption{Normalized Precision Plots on OTB}
    \label{fig-otb-ablation-normed-precision-performance}
  \end{subfigure}
  \caption{Ablation studies on the OTB-99-language dataset using the OPE
    protocol.  Red curve is evaluated on our model conditioned on both NL and
    bounding box, while the blue curve is evaluated on our model conditioned on
    NL only.  Dashed-green curve is evaluated on our model conditioned on both
    NL and bounding box but without relaxed gating during tracking.
    Dashed-magenta curve is evaluated on our NL detection module.}
  \label{fig-ablation-study}
\end{figure*}

\subsection{Datasets and Evaluation Criteria}
Our problem statement of combining NL description and appearance for
tracking~\cite{Li} differs from the majority of prior published
works~\cite{bertinetto2016fully,danelljan2016beyond,song2018vital}.
This opens the possibility for new ways of evaluating trackers but also
presents challenges in comparing to previous trackers.

In particular, these challenges stem from the fact that a video clip can be
labeled with a diverse set of NL descriptions. These
descriptions can range from precise, e.g., ``a sliver sedan'' to vague, e.g.,
``a monkey among other monkeys.'' Handling ambiguous or conflicting
descriptions is outside the scope of this work.

We are aware of two publicly-available tracking datasets cited
in~\cite{fan2018lasot} and~\cite{Li} that have NL annotations
defining the target to track. These datasets exhibit biases in terms of the
linguistic style of the annotations, and how precise they are in defining the
target to be tracked. Nonetheless, we adopt these two datasets,
LaSOT~\cite{fan2018lasot} and otb-99-lang~\cite{Li} for our evaluation to allow
an informative comparison with our baselines.

The problem of language-description bias in these datasets is prominent, both
in how natural the NL description seems to a native English speaker and whether
or not it uniquely defines the target specified by the ground-truth bounding
box.

While the former, i.e., ``naturalness'' of the descriptions is hard to
quantify, ambiguity of the descriptions can be tested. We set up a simple task
where outsourced workers rely on the NL description of each video to select the
object to be tracked for randomly sampled frames. We aggregate videos from
LaSOT that pass this test into a dataset we call ``NL-consistent LaSOT''. Our
crowd workers determine 21 videos from LaSOT testing set are NL-consistent.  We
observe a decisive improvement in the effectiveness of our tracker in handling
occlusions and fast motion in videos from NL-consistent LaSOT compared to
LaSOT.

We adopt standard evaluation metrics for visual object trackers:
1. One pass evaluation (OPE) of success rate at different intersection over
    union (IoU) area under curve (AUC), and
2. OPE of mean average precision (MAP) over location error threshold AUC.

\begin{figure*}
  \centering
  \begin{subfigure}[b]{0.325\textwidth}
    \includegraphics[width=\textwidth]{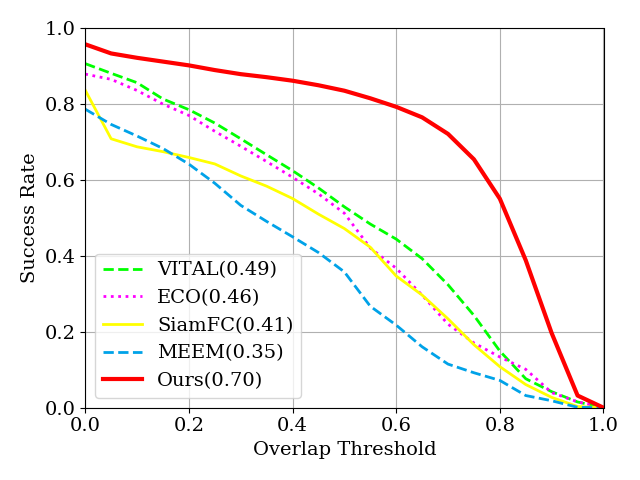}
    \caption{Success Plots}
    \label{fig-lasot-ablation-success-performance}
  \end{subfigure}
  \begin{subfigure}[b]{0.325\textwidth}
    \includegraphics[width=\textwidth]{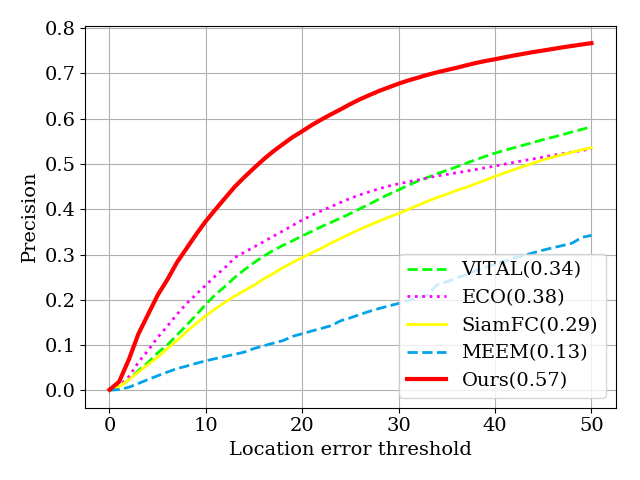}
    \caption{Precision Plots}
    \label{fig-lasot-ablation-precision-performance}
  \end{subfigure}
  \begin{subfigure}[b]{0.325\textwidth}
    \includegraphics[width=\textwidth]{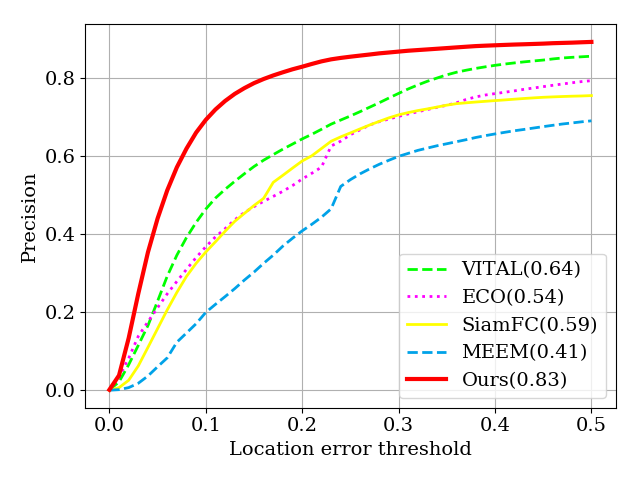}
    \caption{Normalized Precision Plots}
    \label{fig-lasot-ablation-normed-precision-performance}
  \end{subfigure}
  \caption{Performance of trackers on ``NL-consistent LaSOT": one pass
    evaluation success rate at different intersection over union thresholds and
    mean average precision / mean average normalized precision at different
    location error thresholds on testing videos of LaSOT (``NL-consistent
    LaSOT") with NL annotations verified by crowd sourcing workers. Our model is conditioned on NL and bounding box.}
  \label{fig-verified-performance}
\end{figure*}

\begin{figure*}[!t]
  \begin{center}
      \begin{subfigure}[b]{0.19\textwidth}
        \includegraphics[width=\textwidth]{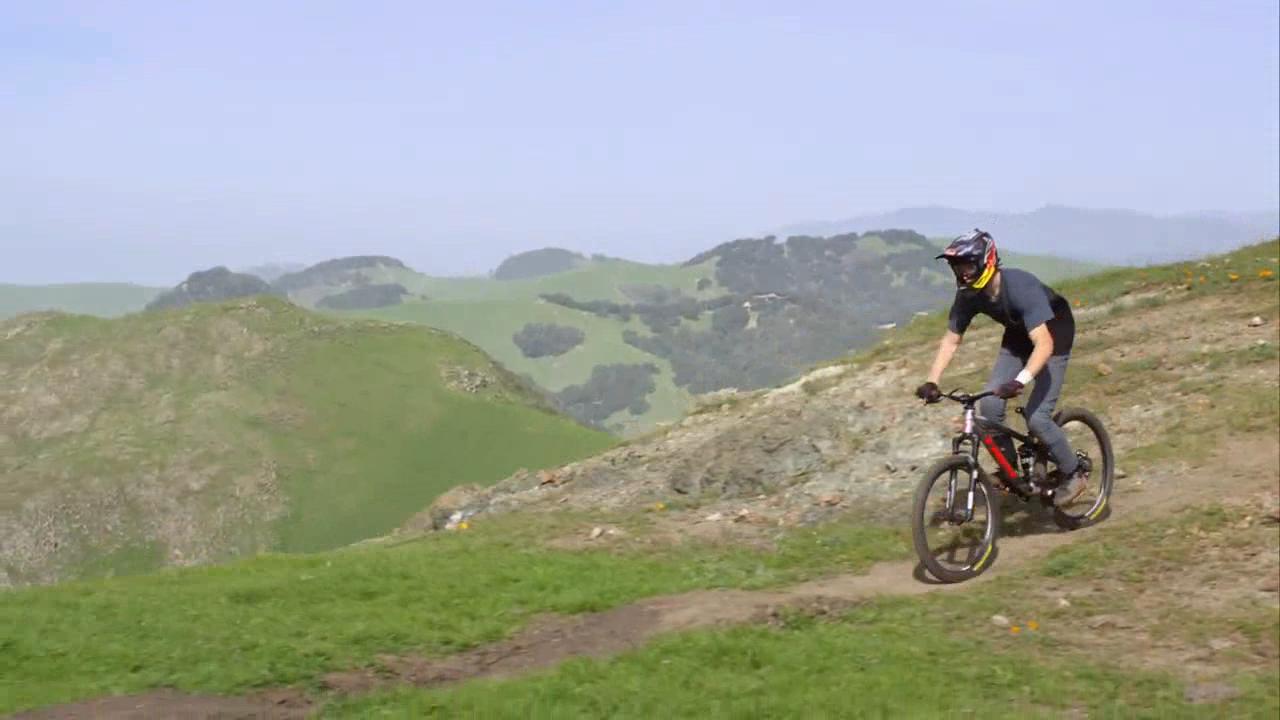}
      \end{subfigure}
      \begin{subfigure}[b]{0.19\textwidth}
        \includegraphics[width=\textwidth]{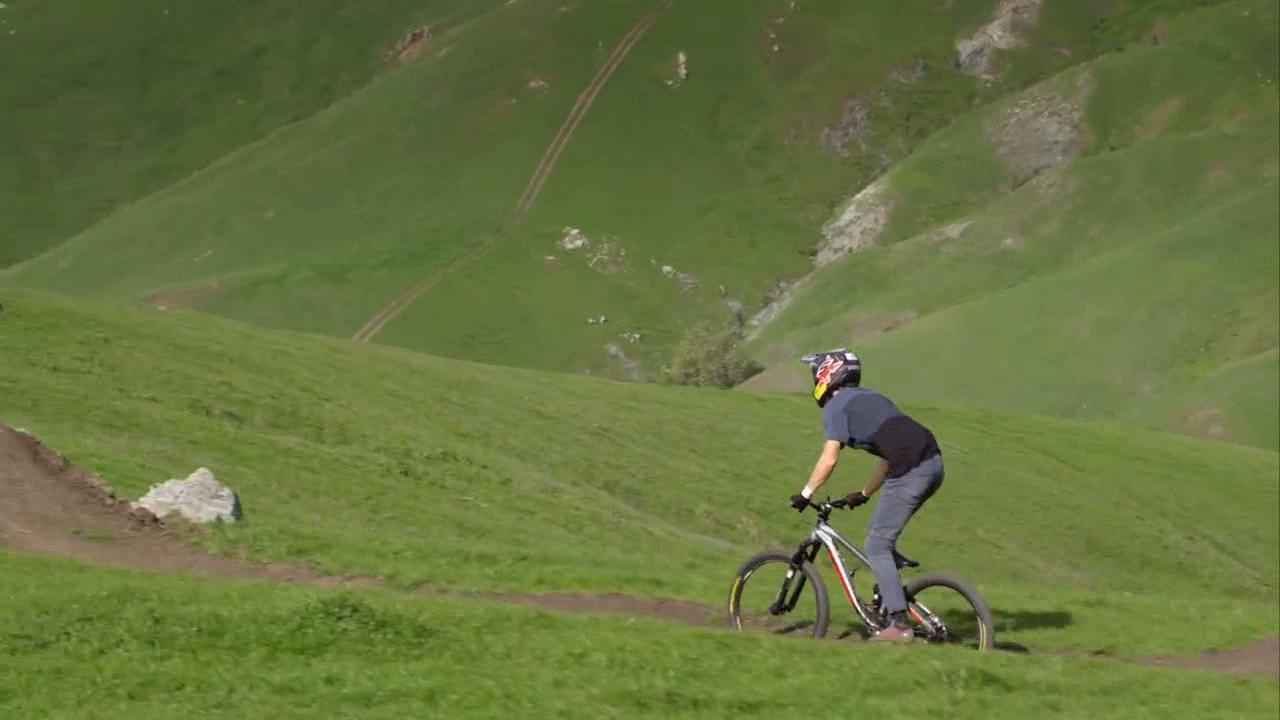}
      \end{subfigure}
      \begin{subfigure}[b]{0.19\textwidth}
        \includegraphics[width=\textwidth]{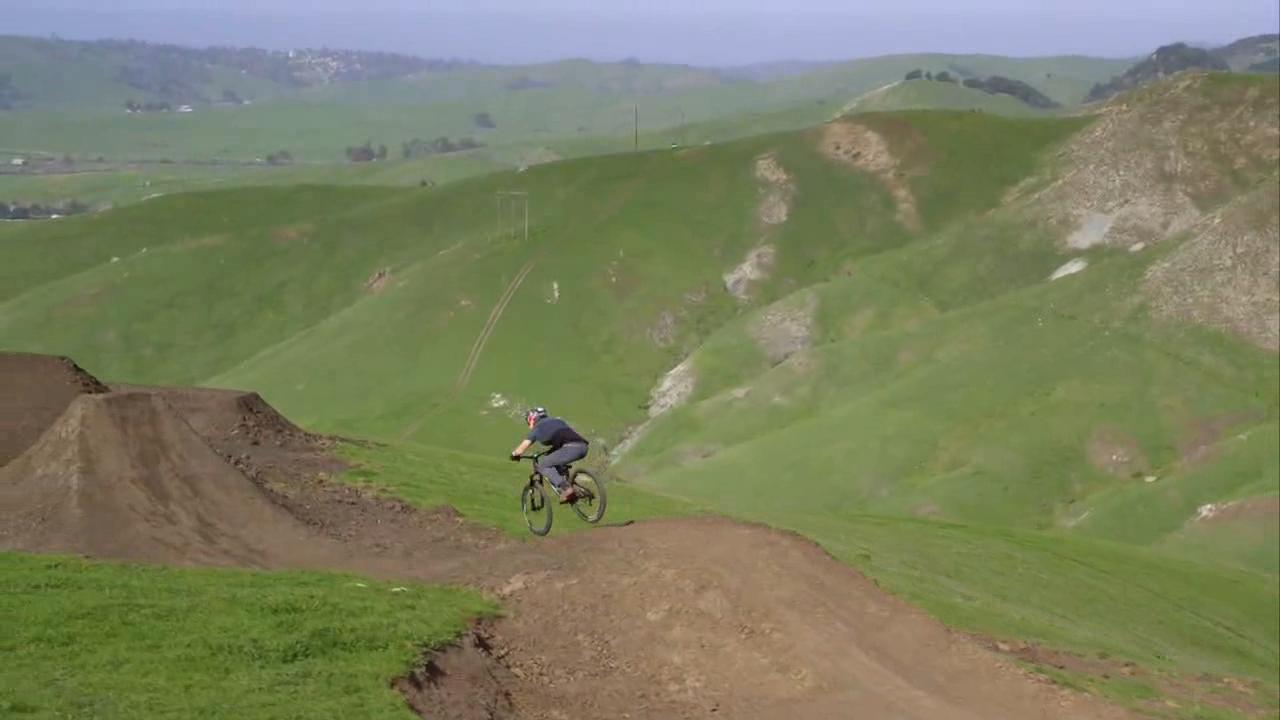}
      \end{subfigure}
      \begin{subfigure}[b]{0.19\textwidth}
        \includegraphics[width=\textwidth]{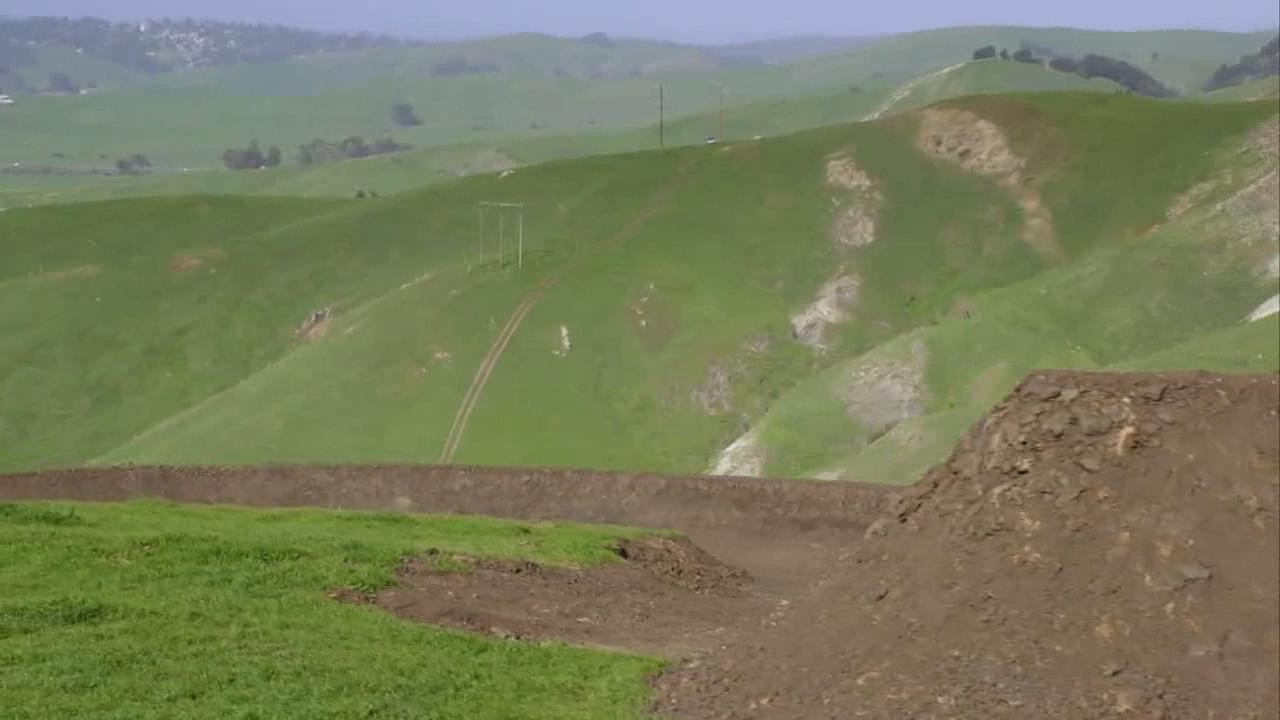}
      \end{subfigure}
      \begin{subfigure}[b]{0.19\textwidth}
        \includegraphics[width=\textwidth]{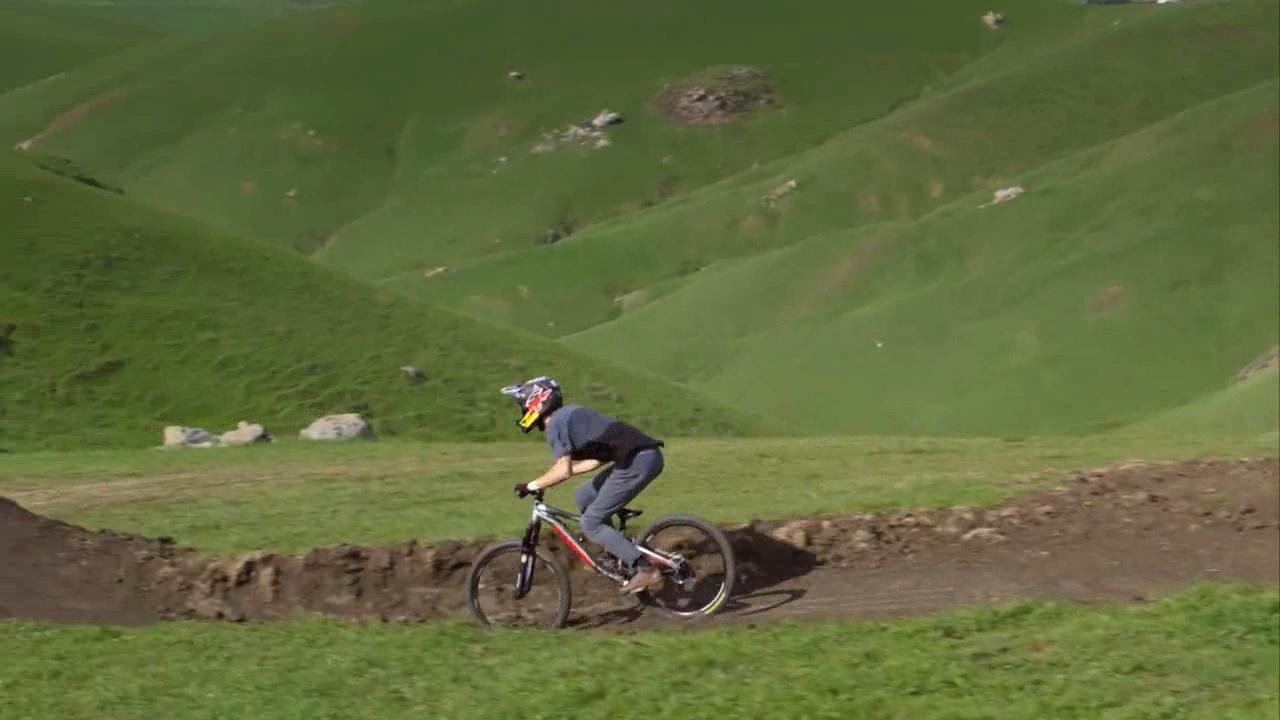}
      \end{subfigure}
    \end{center}
    \vspace{-0.25 in}
\end{figure*}
\setcounter{figure}{\the\numexpr\value{figure}-1\relax}
\begin{figure*}[!t]
  \begin{center}
    \resizebox{2\columnwidth}{!}{
      \begin{tikzpicture}
        \begin{axis}[
            width=2\columnwidth,
            height=0.55\columnwidth,
            xlabel = Frame Num,
            ylabel = IoU,
            legend pos = outer north east 
          ]
          \draw[color=gray!50,fill=gray!50] plot[smooth, samples=20, domain=380:391] (\x,1) |- (380,0) -- cycle;
          \draw[color=gray!50,fill=gray!50] plot[smooth, samples=20, domain=495:503] (\x,1) |- (495,0) -- cycle;
          \draw[color=gray!50,fill=gray!50] plot[smooth, samples=20, domain=651:708] (\x,1) |- (651,0) -- cycle;
          \addplot[smooth, color=red, line width=0.30mm] table[x={FRAME}, y={Ours}]{data/overlap-time.txt};
          \addplot[smooth, color=yellow] table[x={FRAME}, y={ECO}]{data/overlap-time.txt};
          \addplot[smooth, color=green] table[x={FRAME}, y={VITAL}]{data/overlap-time.txt};
          \addplot[smooth, color=blue] table[x={FRAME}, y={SiamFC}]{data/overlap-time.txt};
          \addplot[smooth, color=orange] table[x={FRAME}, y={MEEM}]{data/overlap-time.txt};
          \addplot[smooth, color=violet] table[x={FRAME}, y={ROLO}]{data/overlap-time.txt};
          \legend{Ours, ECO, VITAL, SiamFC, MEEM, ROLO}
        \end{axis}
      \end{tikzpicture}
    }
    \vspace{-0.1 in}
    \caption{IoU vs.~time into tracking on \textit{bicycle-18} from
      LaSOT~\cite{fan2018lasot}. The given NL description is: \textit{black bicycle
      by a man on the mountain road}. Our model here is evaluated without
      relaxed gating. Gray temporal intervals indicate full occlusions. Frames on the top are sampled from the video to show how occlusion happened.} 
    \label{fig-overlap-time}
    \vspace{-0.3 in}
  \end{center}
\end{figure*}
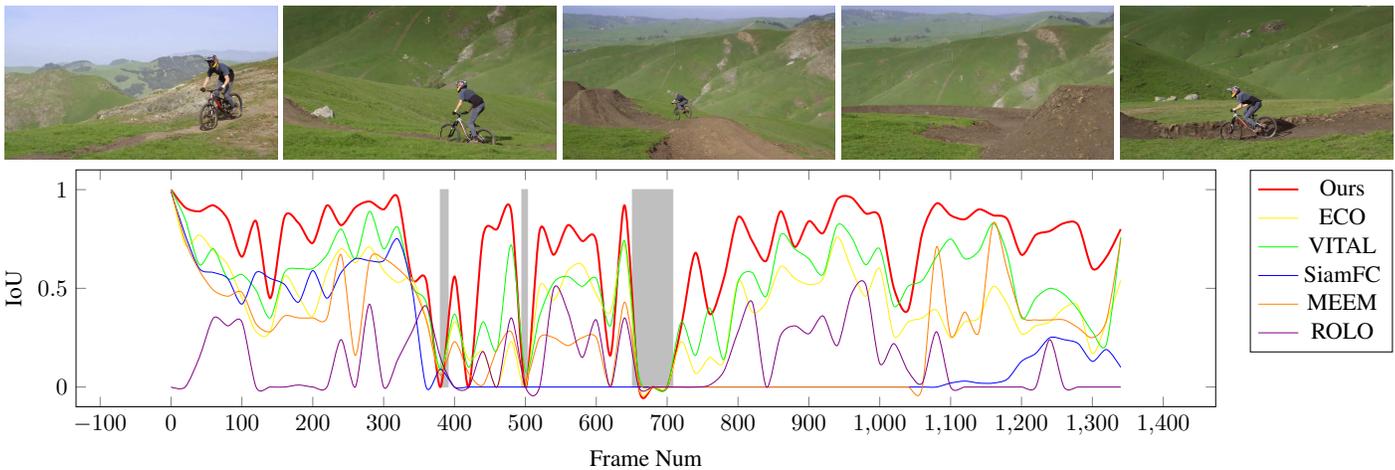

\subsection{Language Trackers}~\label{sec-lang-tracker-performance}
The contribution of this work is a tracker that exploits both visual cues and
NL descriptions of the target. Li et al.~\cite{Li}
introduced the tracking by natural language specification problem.
It is worth noting that initializing a tracker with the bounding box is not free (except when
working on academic datasets) and cannot be taken for granted. Therefore, we compare
the proposed model with \cite{Li} in two different set-ups of the
problem: tracking with or without the initialization of a bounding box.

As no training codes are available for~\cite{Li}, we omit its results on LaSOT.
The results are summarized in Table~\ref{tbl-lang-performance} and
Fig.~\ref{fig-nl-tracker}. These results show that by adopting a
tracking-by-detection framework, when trackers are conditioned on only NL, our
model doubles the tracking performance of the previous best attempt~\cite{Li}.
Additionally, when a ground truth bounding box is jointly provided with the NL
description, our model outperforms Li and others' work~\cite{Li}.

\subsection{Comparison with Visual Object Trackers}~\label{sec-performance}
We follow the standard evaluation procedures of OPE of success rate over IoU
curve and MAP over location error threshold curve, and we compare our proposed
tracker with the following trackers: Li and others' Model I~\cite{Li},
VITAL~\cite{song2018vital}, SiamFC~\cite{bertinetto2016fully},
ECO~\cite{danelljan2017eco}, and MEEM~\cite{zhang2014meem}.  The results are
summarized in Fig.~\ref{fig-performance}. Our tracker is competitive with state
of the art trackers.

On ``NL-consistent LaSOT" our tracker out-performs all other trackers by 20
percentage points on AUC of the success plots in
Fig.~\ref{fig-verified-performance}.  This is due to the verified NL
description which makes our tracker better at detection and hence yields a
better tracking performance.

\textbf{Mean IoU over Time: }
In addition to these standard evaluation metrics, we also qualitatively evaluate
the performance of trackers based on the OPE mean IoU over time curve.
For long term tracking, the inability to re-acquire the target after losing the
target during tracking is fatal to trackers. The mean IoU over time curve gives
us a qualitative way of evaluating trackers in terms of long-term tracking.  We
plot the IoU for test videos at each time step, in which the horizontal axis is
the index of frame and the vertical axis is IoU.

As our proposed tracker employs tracking by detection, language based detection
plays an important role in the entire tracking process. An example of IoU at
different time steps is shown in Fig.~\ref{fig-overlap-time}. It shows that our
tracker can re-acquire the target with NL descriptions after losing the target
due to cases like occlusion and/or rapid motion.

\subsection{Ablation Studies}~\label{sec-ablation-studies}
In this section, we study the contribution of key aspects of our formulation.
We focus on the OTB-99-lang dataset and follow the OPE protocol to enable a
more comprehensive comparison with previously-published methods. The results of
this ablation study are summarized in Fig.~\ref{fig-ablation-study}.

In Fig.~\ref{fig-otb-ablation-success-performance}, we observe that disabling
relaxed-gating has a relatively minor effect on performance, dropping
from 0.61 to 0.6 in terms of success rate. However, the red and dashed-green
curves reveal a more complex story: gating-free tracking appears to yield
a higher success rate in the low-accuracy regime, but under-performs badly as
we ``require'' higher accuracy.  This is not surprising: accepting temporally
implausible language proposals tends to yield poor tracking results.
Conditioning our tracker solely on NL (the blue curve) does not have a
substantial effect up to the threshold of 0.5. However, the red curve dominates
the blue when using higher thresholds of IoU for evaluation.

Recall from Fig.~\ref{fig-otb-success-performance} that our pure-NL tracker is
competitive with SiamFC~\cite{bertinetto2016fully}. The remaining curve,
dashed-magenta corresponds to picking one bounding box per frame, the one that
scores highest by our NL detection phase. Not surprisingly, such a
``tracker'' does not fare well, underscoring the benefit of our LSTM-based
formulation.

In summary, our ablation study supports our statement about the benefit of NL.
It's worth noting that this benefit can only be obtained by deriving a correct
tracking formulation.
\section{Conclusion}
We have presented a novel approach to condition a visual tracker on a natural
language description of its target. Unlike prior attempts, we propose a novel
track-by-detection route: our formulation employs a deep network to propose
image regions that are most likely given the NL description and an RNN that
implicitly maintains the history of appearance and dynamics. Experiments on
challenging datasets demonstrate that our NL tracker is competitive with the
state of the art using standard performance metrics. More importantly, our
tracker enjoys better stability in terms of re-acquiring the target after rapid
motion or occlusions than conventional trackers. This property supports our
claim that NL descriptions provide additional invariants for the target (i.e.,
beyond appearance) and if exploited correctly, can yield a tracker that is more
robust and stable to occlusions.

{\small
  \bibliographystyle{ieee}
  \bibliography{bib.bib}
}
\end{document}